\documentclass[journal]{IEEEtran}

\usepackage{amsmath,amssymb,amsfonts}
\usepackage[pdftex]{graphicx}
\usepackage[linesnumbered,ruled,vlined]{algorithm2e}
\usepackage{subfigure}
\usepackage{booktabs}
\usepackage{diagbox}

\ifCLASSINFOpdf
\else
\fi
\begin{document}
%
\title{A Lightweight Privacy-Preserving Scheme Using Label-based Pixel Block Mixing for Image Classification in Deep Learning}
%
%
%

\author{Yuexin~Xiang\thanks{The first two authors contributed equally to this work.},~\IEEEmembership{Student Member,~IEEE,}
        Tiantian~Li,
        Wei~Ren\thanks{Corresponding Author: Wei Ren (E-mail:weirencs@cug.edu.cn).},~\IEEEmembership{Member,~IEEE,}
        Tianqing~Zhu,~\IEEEmembership{Member,~IEEE,}
        and~Kim-Kwang~Raymond~Choo,~\IEEEmembership{Senior~Member,~IEEE}
\thanks{Y. Xiang is with the School of Computer Science, China University of Geosciences, Wuhan, P.R. China.}
\thanks{T. Li is with the School of Computer Science, China University of Geosciences, Wuhan, P.R. China.}
\thanks{W. Ren is with the School of Computer Science, China University of Geosciences, Wuhan, P.R. China; Henan Key Laboratory of Network Cryptography Technology, Zhengzhou, P.R.  China; Key Laboratory of Network Assessment Technology,
CAS, Institute of Information Engineering, Chinese Academy
of Sciences, Beijing 100093, P.R. China}
\thanks{T. Zhu is with the School of Computer Science, China University of Geosciences, Wuhan, P.R. China.}
\thanks{K.-K.R. Choo is with the Department of Information Systems
and Cyber Security, University of Texas at San Antonio,
San Antonio, TX 78249-0631, USA.}
}

\maketitle

\begin{abstract}
To ensure the privacy of sensitive data used in the training of deep learning models, a number of privacy-preserving methods have been designed by the research community. However, existing schemes are generally designed to work with textual data, or are not efficient when a large number of images is used for training. Hence, in this paper we propose a lightweight and efficient approach to preserve image privacy while maintaining the availability of the training set. Specifically, we design the pixel block mixing algorithm for image classification privacy preservation in deep learning. To evaluate its utility, we use the mixed training set to train the ResNet50, VGG16, InceptionV3 and DenseNet121 models on the WIKI dataset and the CNBC face dataset. Experimental findings on the testing set show that our scheme preserves image privacy while maintaining the availability of the training set in the deep learning models. Additionally, the experimental results demonstrate that we achieve good performance for the VGG16 model on the WIKI dataset and both ResNet50 and DenseNet121 on the CNBC dataset. The pixel block algorithm achieves fairly high efficiency in the mixing of the images, and it is computationally challenging for the attackers to restore the mixed training set to the original training set. Moreover, data augmentation can be applied to the mixed training set to improve the training's effectiveness.
\end{abstract}


%
\IEEEpeerreviewmaketitle

\section{Introduction}
%
%
%
%
\IEEEPARstart{D}{igitalization} of our society has resulted in significant data being sensed, generated, shared, and processed (e.g., by devices such as the Internet of Things devices and systems, and user-generated content), and such data can be used to train deep learning models in various applications ranging from smart buildings to smart cities, and so on. As technologies advance, so does our understanding and expectation of privacy. For example, data used in the training of deep learning models can contain user-sensitive information such as our facial images, as explained in a recent \emph{Nature} article \cite{van2020ethical}. This reinforces the importance of protecting the private data in the training set while maintaining the utility of the trained deep learning models. 

A number of privacy-preserving approaches have been proposed in the literature, such as using differential privacy techniques in training deep learning models \cite{abadi2016deep,mcmahan2017learning,ryffel2018generic,carlini2019secret}, using homomorphic encryption in deep learning model training \cite{nikolaenko2013privacy, yuan2013privacy, bost2015machine,takabi2016privacy, li2017multi, aono2017privacy,hesamifard2018privacy}, and various other techniques \cite{shokri2015privacy, xu2015privacy, wang2018not, ma2018privacy}. Most of these existing approaches are generally designed to work with textual information, and may not be applicable for (large) image privacy preservation application. 

Several image privacy preservation schemes, such as those presented in \cite{hsu2011homomorphic,qin2014towards,maekawa2018privacy,he2016puppies,li2019privacy,ra2013p3,zhang2015pop}, have been proposed although they are not designed to preserve privacy during the deep learning training process. Hence, approaches to ensure image privacy preservation during the training of deep learning models have also been designed \cite{tanaka2018learnable,sirichotedumrong2019privacy,li2019deepobfuscator,wu2019p3sgd}. However, achieving a balance between efficiency and privacy remains challenging to achieve.

In this paper, we propose a novel approach based on the pixel block mixing algorithm to prevent image data privacy leakage in deep learning training. We then evaluate the effectiveness of our scheme by comparing the training results from the original training set and the mixed training set from the same dataset with the same deep learning model structures and parameters. We also apply data augmentation to the mixed training set, transformed by the pixel block mixing algorithm, to improve the training effect for the deep learning models. In the next section, we will review the related literature.

\section{Related Work}
\label{sec:2}

\subsection{Differential Privacy and Homomorphic Encryption}

Designing robust and privacy-preserving deep learning models, such as generative text models, can be challenging. As discussed earlier, differential privacy is one of the approaches used to minimize the risk of data reconstruction and model memorization. For example, Abadi et al. \cite{abadi2016deep} proposed a novel differential privacy framework to control the privacy budget during the training of deep neural networks, in order to achieve the required training effect and improve the protection of data privacy. McMahan et al. \cite{mcmahan2017learning} proposed training recurrent language models with user-level differential privacy guarantees, but computation overhead in their approach is relatively high. Ryffel et al. \cite{ryffel2018generic} proposed a framework that uses both secure multiparty computation and differential privacy to establish complex privacy-preserving structures in deep neural networks. 

Privacy-preserving approaches based on homomorphic encryption include those presented in \cite{yuan2013privacy}. Specifically, the authors of \cite{yuan2013privacy} presented a scheme in which participants upload their homomorphic encrypted private data to the cloud for the training of the neural network. However, this method supports only a small number of participants. Bost et al. \cite{bost2015machine} proposed a privacy-preserving classification protocol, based on homomorphic encryption, to classify encrypted data. However, this model is slow and works well for only small datasets. Li et al. \cite{li2017multi} presented two multi-key privacy-preserving deep learning schemes that are based on fully homomorphic encryption. They proved that both schemes are secure. Hesamifard et al. \cite{hesamifard2018privacy} proposed a scheme to apply the deep neural network to the encryption domain, and also presented a method to enhance the deep neural network in homomorphic encryption for protecting users' private raw data. 

\subsection{Image Privacy Preservation in Deep Learning}

Differential privacy- and homomorphic encryption-based approaches are hard to apply in image data on deep learning models, because of the characteristics of the algorithms and the effect of the data type(s). Tanaka \cite{tanaka2018learnable} proposed encrypting the image, so that the human can not understand the encrypted image, but the network can be trained with the encrypted image. The scheme allows us to train the network without the associated privacy implication. Also, a new pixel-based image encryption method for deep neural networks for privacy preservation was proposed by Sirichotedumrong et al. \cite{sirichotedumrong2019privacy}. In addition, they considered a new adaptive network that reduces the impact of image encryption. Li et al. \cite{li2019deepobfuscator} proposed the adversarial training framework DeepObfuscator, which can prevent attackers from reconstructing the original image or inferring private attributes, while retaining information required for image classification. Wu et al. \cite{wu2019p3sgd} introduced a novel stochastic gradient descent (SGD) protocol, which performs a step-by-step update of the patient-level SGD model for protecting private images from the patients. Specifically, they proposed to inject updates with carefully designed noise to achieve privacy-preserving and normalizing of the convolutional neural network model. They also equipped their scheme with a well-designed strategy to control the scale of injected noise adaptively.

\subsection{Summary}
There are a number of limitations in existing approaches, for example how to strike a balance between privacy preservation and efficiency during the training of deep learning models using a large number of images. For example, using encryption operations to preserve privacy is computationally expensive when the image data size is large and consequently, the approaches will be inefficient. Approaches, such as that of Wu et al.  \cite{wu2019p3sgd}, with too many steps are also not efficient. Hence, we consider designing a simple and efficient method to preserve the privacy of image data while maintaining the availability of the training set for training the deep learning models. According to the process of image recognition, the deep learning model continuously extracts the features in the image. Therefore, we posit that it is possible to train a usable deep learning model by processing images with the same label in the training set in a certain regular way. This motivates the design of our proposed lightweight and flexible pixel block mixing algorithm to preserve the privacy of each image in the training set.

\section{Problem Formulation}
\label{sec:pf}
In this section, we will illustrate the adversarial model and the definition of privacy respectively.

\subsection{Adversarial Model}

Mireshghallah et al.\cite{mirshghallah2020privacy} divide threats against deep learning into two types: direct information exposure and indirect (inferred) information exposure. 

In this paper, we mainly examine direct information exposure. In this kind of attack, access to the actual data can be obtained by the attackers. In the direct information exposure threat, there are two main situations, which may happen during the training process or non-training process: 
\begin{itemize}
    \item The hackers access the sensitive training set by the malware, social engineering, and other methods without authorization.
    \item The administrators or entities who manage the data of the sensitive training set leak the data unintentionally or purposely.
\end{itemize}

Consequently, our research concentrates on preserving the privacy 
of the training set consists of image type data, especially human face images.

\subsection{Definition of Privacy}
The definition of privacy in this paper is the special characteristic information combination of the image. Furthermore, the face images have special great value due to the rapid development of face recognition is benefited from the deep learning technology. The training of deep learning models for recognizing human faces requires a large number of individual face images. 

Besides, because the human face images include sensitive private information of people, the revealed personal characteristic information combinations may be illegally used by the attackers, such as identity theft and face recognition, which damages the personal information security and possibly cause greater losses on finance or other losses. 


In order to preserve the privacy, we assume the image set $A$ has $n$ types of image $A = \{A_1,A_2,...,A_n\}$. $A_i$ is the image type set in $A$, $a_i$ is the element in $A_i$, and $a_x$ is a randomly selected image. In order to preserve the image's privacy while maintaining the utility of the training set, it should meet the conditions below:
\begin{equation} \label{privacy}
a_x \in A_i, a_x \nRightarrow a_i
\end{equation}

Specifically, the core idea of Eq.(\ref{privacy}) is randomly mixing the images with the same label that can protect the personal characteristic information combination. Also, we tested our scheme on the human face datasets. Then, we will describe the particular scheme in the following section.

\section{Proposed Scheme}
\label{sec:3}
A summary of notation used in this paper is presented in Table \ref{tab-nota}.

\begin{table}[!htbp]
\setlength{\belowcaptionskip}{+0.1cm}
\caption{Notations} \label{tab-nota} \centering
\begin{tabular}{l l}
\noalign{\hrule height 0.5pt}
Notations & Description\\
\noalign{\hrule height 0.5pt}
$E$ & the matrix\\ 
$D$ & the element of the partitioned matrix\\
$d$ & the element of the matrix\\
$I_S$ & the image dataset\\
$I_T$ & the mixed image dataset\\ 
$I_i$ & the image in $I_S$\\
$I_l$ & the image selected from $I_S$ that have the same\\
$ $ & label as the targeted image\\
$N_{I_{S}}$ & the number of $I_i$s in $I_S$\\
$N_{s}$ & the number of $I_l$s\\
$N_b$ & the number of divided pixel blocks\\
$N_t$ & the number of times of pixel block mixing per\\
$ $ & image in $I_S$\\
$N_{p}$ & the total number of pixels in $I_i$\\
$L_b$ & the length of each divided pixel block\\
$L_{i}$ & the length of $I_i$\\
$W_b$ & the width of each divided pixel block\\
$W_{i}$ & the width of $I_i$\\
$S_{01}$ & the random 0-1 sequence\\
$B_{I_{i}}$ & the divided pixel blocks of $I_i$\\
$B_{I_{l}}$ & the divided pixel blocks of $I_l$\\
$b_{i}$ & the pixel block of $B_{I_{i}}$\\
$b_{l}$ & the pixel block of $B_{I_{l}}$\\
\noalign{\hrule height 0.5pt}
\end{tabular}
\end{table}

\subsection{Theoretical Model}
We choose an image from a dataset with the same label which is equal to a matrix $E$ with the same size. Thus, for mixing the features in the dataset while keeping parts of features unchanged, first we consider converting $E$ to the partitioned matrix as shown below:
\begin{equation}\label{m_main}
E =
\left[
\begin{array}{ccc}
    d_{11} & \cdots & d_{1n} \\
    \vdots & \ddots & \vdots \\
    d_{m1} & \cdots & d_{mn} 
\end{array}
\right]
\Rightarrow
\left[
\begin{array}{ccc}
    D_{11} & \cdots & D_{1n'} \\
    \vdots & \ddots & \vdots \\
    D_{m'1} & \cdots & D_{m'n'} 
\end{array}
\right]
\end{equation}

Also, in our scheme, the numbers of rows and columns are the same for different $D_{ij}$ in the partitioned matrix (i.e. we divide the matrix evenly into the partitioned matrix). In order to achieve Eq.(\ref{privacy}), other images in the dataset that have the same labels should be divided the same as the method in Eq.(\ref{m_main}). We assume that the partitioned matrices corresponding to other images are expressed in the following form:
\begin{equation}\label{m}
E_{k} =
\left[
\begin{array}{ccc}
    D_{k_{11}} & \cdots & D_{k_{1n'}} \\
    \vdots & \ddots & \vdots \\
    D_{k_{m'1}} & \cdots & D_{k_{m'n'}} 
\end{array}
\right]
, k \in \mathbb{N}
\end{equation}

Next, we mark the event $D_{ij}$ being substituted by $D_{k_{ij}}$ as $D_S$ and its probability as $P_D$, that is: 
\begin{equation}\label{prob}
P(D_{ij} \leftarrow D_{k_{ij}}) = P(D_S) = P_D
\end{equation}
Besides, we select $N_E$ matrices to participate the mixing. For minimizing the influence on the features of the original data set, we only replace the blocks of $E$ to the blocks in other matrices $E_k$ that are at the same location as shown in Eq.(\ref{prob}). 

Finally, for each $E_K$ in selected $N_E$ matrices, we will execute the following operation for each block in $E$ according to $P_D$ mentioned in Eq.(\ref{prob}), then operating the same process on each $E$ in the dataset:
\begin{equation}\label{achieve}
\left\{
\begin{aligned}
D_{ij} \leftarrow D_{k_{ij}}, D_S = true \\
D_{ij} \leftarrow D_{ij}, D_S = false
\end{aligned}
\right.
\end{equation}

\begin{figure}[!htbp]
  \centering
  \includegraphics[width=0.6\linewidth]{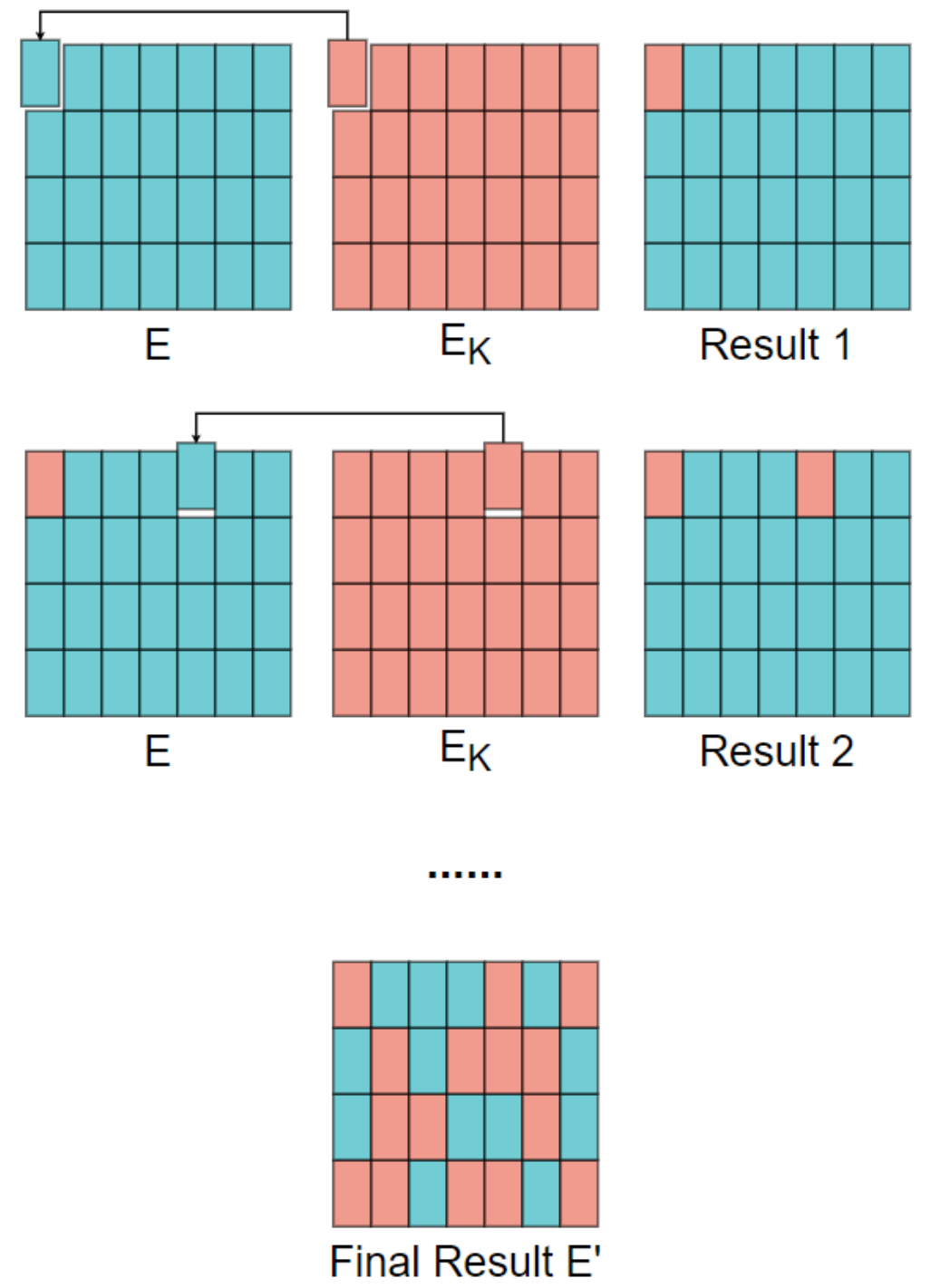}
  \caption{The process and result of the mixing}
  \label{m_show}
\end{figure}

After applying Eq.(\ref{achieve}), the new generated $E'$ will look like the matrix mixed by $E$ and $E_K$ shown in Fig.{\ref{m_show}}.

In summary, Eq.(\ref{achieve}) is the method we achieve the goal proposed in Eq.(\ref{privacy}). In the next part, we will illustrate the implement of the theoretical model. 

\subsection{Scheme Implement}

We will now describe the implement of the pixel block mixing algorithm based on our proposed theoretical model. Prior to describing the algorithm, we will set the algorithm's parameters (i.e., $N_t$, $N_{s}$, $L_b$ and $W_b$). To compute $N_b$:
\begin{equation}
N_b = N_{p} / (L_b \cdot W_b)
\end{equation}

Furthermore, $L_b$ and $W_b$ are used to indirectly set the shape and the number of divided blocks. Subsequently, we apply the pixel block mixing algorithm to a selected $I_S$. When $N_t = 1$, the steps are listed as follows:
\begin{itemize}
    \item Traverse each $I_i$ in $I_S$.
    \item Select $N_{s}$ $I_l$s for each $I_i$ in $I_S$ randomly.
    \item Divide $I_i$ into $N_b$ blocks according to $L_b$ and $W_b$.
    \item Generate a $S_{01}$ of length $N_b$ that each $b_i$ corresponds to a value in $S_{01}$.
    \item Traverse each $b_i$ in $B_{I_{i}}$ to generate ${I_i}^{'}$:
    \begin{itemize}
        \item If $b_i \Rightarrow S_{01} = 1$: Randomly select a $I_l$ from $I_l$s and replace $b_i$ by $b_l$ that is in the same location in selected $I_l$; OR
        \item If $b_i \Rightarrow S_{01} = 0$: Keep $b_i$ unchanged.
    \end{itemize}
    \item Form a new set $I_T$ that consists of each ${I_i}^{'}$.
\end{itemize}

For the penultimate step, each $b_i$ in an $I_i$ of $I_S$ has a 50\% probability of being replaced in each round of mixing. Moreover, we can adjust the generation method of the random sequence and the corresponding function relationship to adjust the probability of each $b_i$ being replaced (e.g. generate a random sequence with 0, 1 and 2). The completed pixel block mixing algorithm is shown in Algorithm \ref{alg1}. Also, the code of Algorithm \ref{alg1} has been uploaded to our GitHub\footnote{https://github.com/oopshell/Pixel-Blocks-Mixing-For-Image-Privacy-\\Preservation-In-Deep-Learning}.


\begin{algorithm}
\DontPrintSemicolon
\label{alg1}
\KwIn{$I_S$, $N_t$, $N_{I_{S}}$, $I_i$, $L_{i}$, $W_{i}$, $L_b$, $W_b$, $N_b$, $N_{s}$, $B_{I_{i}}$, $B_{I_{l}}$}
\KwOut{$I_T$}
set $N_t$, $N_{s}$, $L_b$, $W_b$ and select $I_S$;\\
\For{$j$ in range (0, $N_t$)}{ 
    \For{$i$ in range (0, $N_{I_{S}}$)}{
        read $I_i$ as $img$;\\
        resize $img$ to $L_{i}$ $\cdot$ $W_{i}$;\\
        $r$ $\leftarrow$ ($W_{i}$ $/$ $W_b$);\\
        $c$ $\leftarrow$ ($L_{i}$ $/$ $L_b$);\\
        randomly select $N_s$ $I_l$s;\\
        \For{$k$ in range (0, $r$)}{
            generate a $S_{01}$ of length $N_b$;\\
            \For{$m$ in range (0, $c$)}{
                \If{$S_{01}[m]$ $\neq$ 0}{
                    randomly select an integer $R$, $R \in (0, N_{s}]$;\\
                    select the $R^{th}$ $I_l$;\\
                    $B_{I_{i}}[k,m] \leftarrow B_{I_{l}}[k,m]$;\\
                }
                \Else {continue;}
                $m \leftarrow m + 1$;\\
            }
            $k \leftarrow k + 1$;\\
        }
        save $img$ as $img_t$;\\
        $i \leftarrow i + 1$;\\
    }
    $j \leftarrow j + 1$;\\
}
$I_T$ $\leftarrow$ all $img_t$s;\\
\caption{Pixel Block Mixing}
\end{algorithm}

\begin{figure}[ht]
  \centering
  \includegraphics[width=\linewidth]{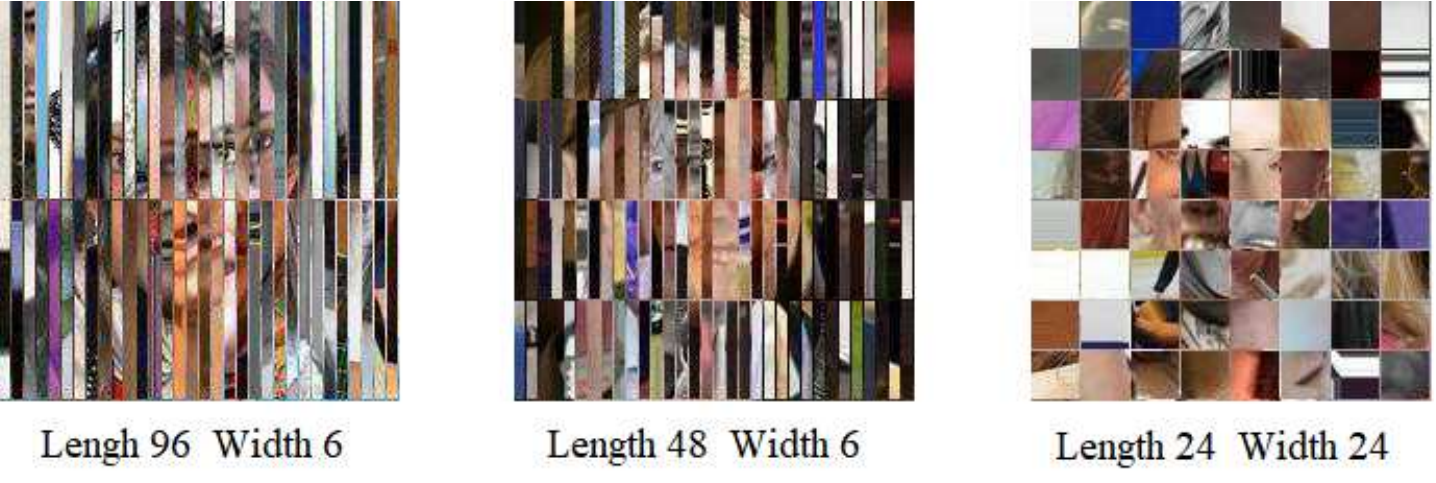}
  \caption{Partial results on the WIKI dataset}
  \label{result}
\end{figure}

The partial results after being processed by the pixel block mixing algorithm on the WIKI dataset are shown in Fig. \ref{result}. From the figure, we observe that it is hard to use those mixed features by a human vision system to determine the individual's identity. 

We will use the new set ${I_S}^{'}$ to train the deep learning models and compare their effectiveness with the models training by the original set $I_S$. The experimental process and results will be explained in the next section.

\section{Experimental Results} \label{sec:4}

\subsection{Basic Settings and Parameter Initialization}
Generally, it is safe to assume that human face images have more exacting privacy requirements than many other textual or image data. Thus, the evaluations will primarily focus on evaluating the algorithm using human face datasets; hence, the choice of both WIKI\footnote{https://data.vision.ee.ethz.ch/cvl/rrothe/imdb-wiki} and CNBC face datasets\footnote{https://wiki.cnbc.cmu.edu/Face\_Place}. 

We run our pixel block mixing algorithm on the original training set $I_S$ to generate a new mixed training set, $I_{T}$. We select ResNet50, VGG16, InceptionV3 and DenseNet121 models for testing. Specifically, we train the deep learning models, by $I_{T}$ with different parameters of image divisions, and compare the results of the models that use the same parameters  (including $epoch$, $batch size$, and $optimizer$) for the same model trained by $I_S$. The images used in testing sets are all normal images.

We initiate the parameters for the experiments on the WIKI dataset and the CNBC face dataset as below:
\begin{equation}
N_{s} = w, N_t = n
\end{equation}
where $w = 10$ on both WIKI and CNBC datasets, $n = 3$ on the WIKI dataset, and $n = 1$ on the CNBC dataset.

Furthermore, to reduce experimental errors, we train each selected classic deep learning model by the same $I_{T}$ for five times on both WIKI  and CNBC datasets and take the average accuracy and loss as the results. Our training experiment environment is the TensorFlow backend-based Keras framework under GPU GTX 1080 Ti.

\subsection{WIKI Dataset Experimental Results}

We use the WIKI dataset as a binary classification dataset for the experiment, and we select gender as the classification criterion. In general, we test ResNet50, VGG16 and InceptionV3 on the WIKI dataset and analyze the trend of the accuracy and the loss. In addition, we record and analyze the accuracy of these three models on the testing set, by mean training the models respectively through the mixed training sets with different division parameters (i.e. $L_b$ and $W_b$). 

\begin{figure}
  \centering
  \includegraphics[width=0.9\linewidth]{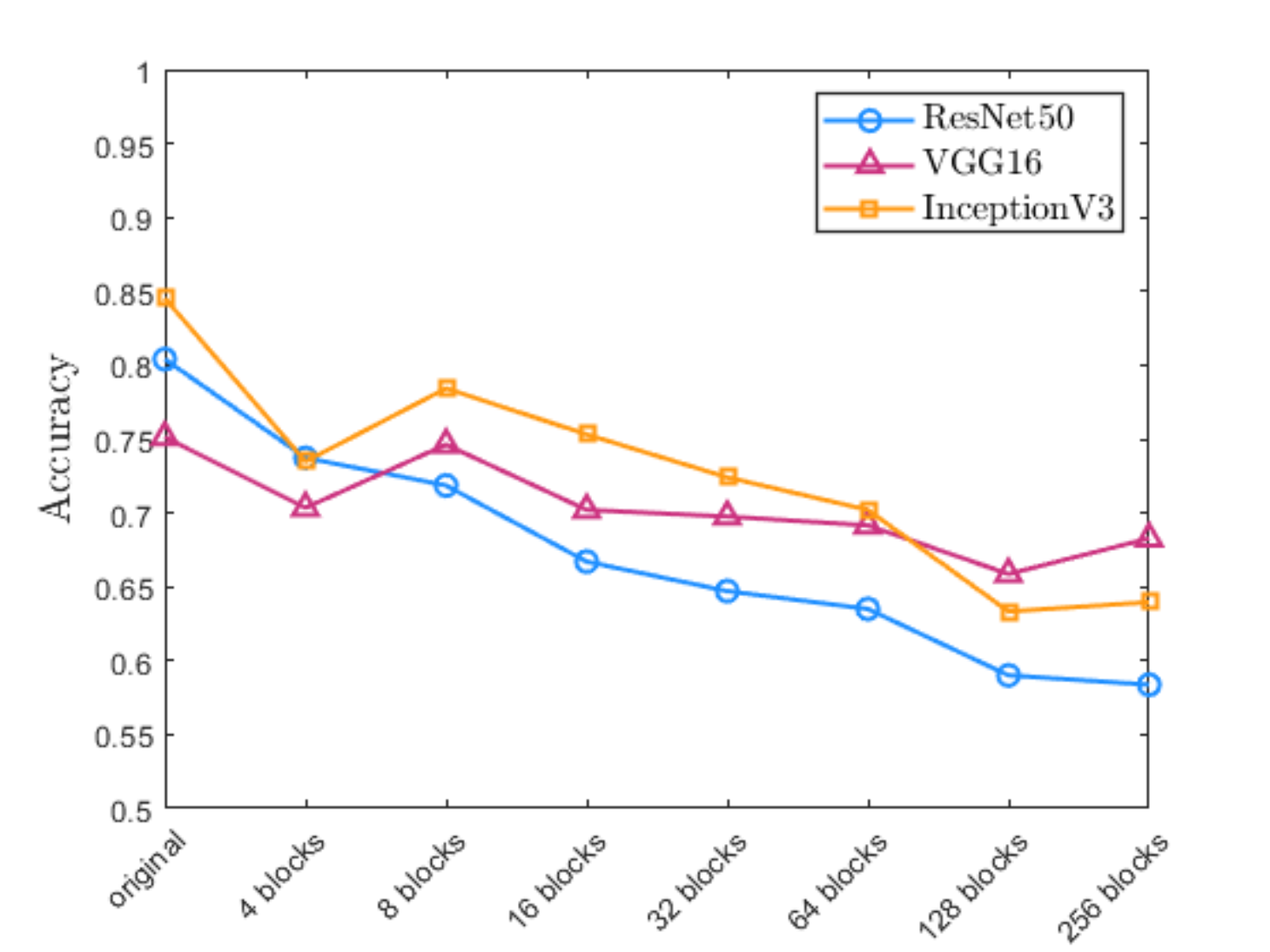}
  \caption{Accuracy achieved on the WIKI dataset}
  \label{WIKItrand}
\end{figure}

From Fig. \ref{WIKItrand}, we observe that as the number of the blocks increases, the accuracy obtained by training ResNet50 and InceptionV3 models progressively decreases but remain at an acceptable level. The VGG16 model remains relatively stable.

\begin{figure}
  \centering
  \includegraphics[width=0.9\linewidth]{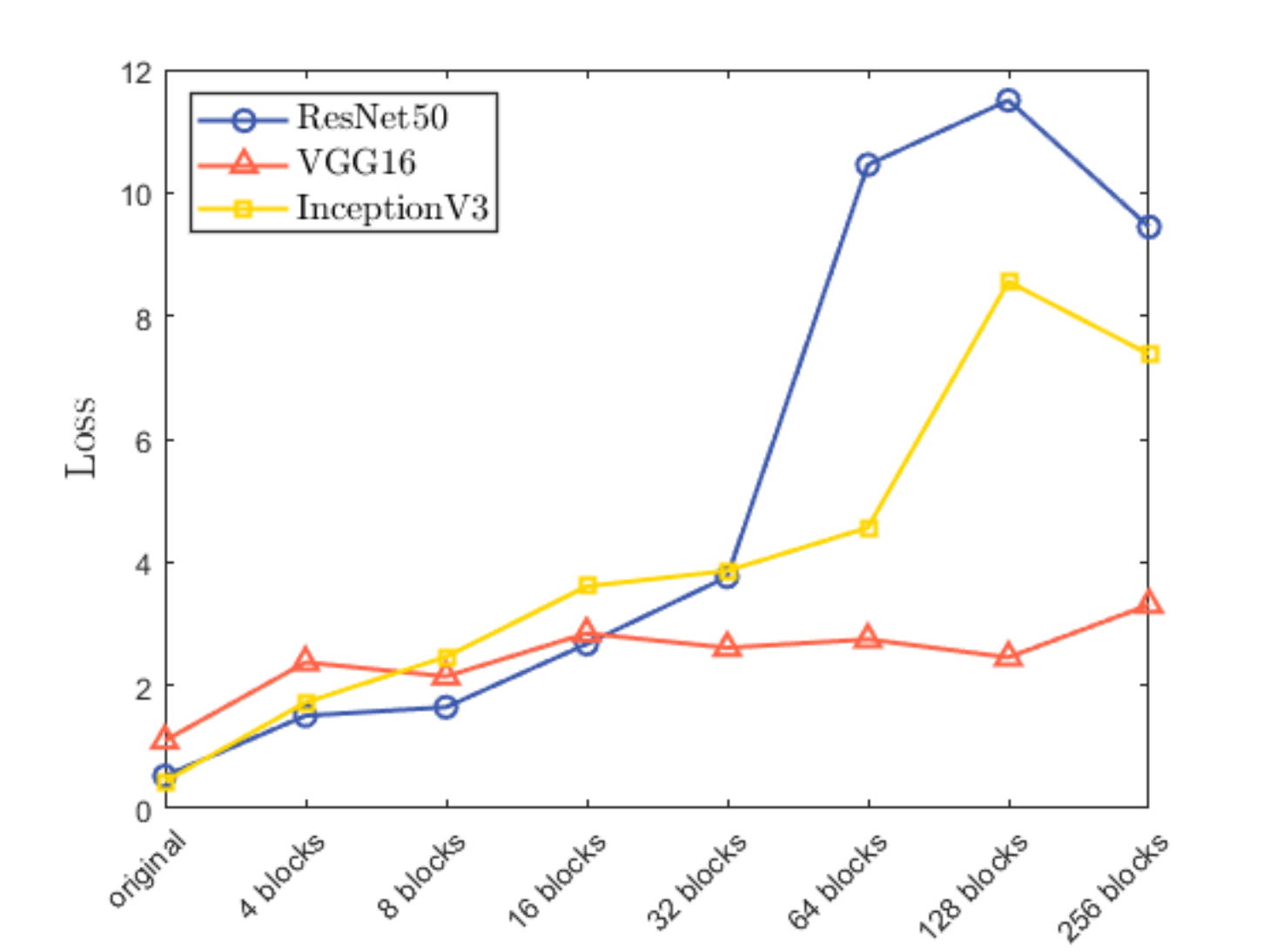}
  \caption{Loss observed on the WIKI dataset}
  \label{WIKIloss}
\end{figure}

Moreover, for Fig. \ref{WIKIloss}, as the number of the blocks increases, the overall loss increases for both ResNet50 and InceptionV3 models, but there is slightly less increase for the VGG16 model. Both ResNet50 and InceptionV3 models experienced significant growth in loss when the number of blocks exceeds 32 and 64, respectively. 

Next, we will explore the differences of the results on the testing set between the accuracy of these deep learning models trained by $I_T$ generated with different division parameters and the accuracy of deep learning models trained by $I_S$. We group $I_T$ with the division parameters of the same $L_b$ together for comparison. The results are shown as follows:


\begin{itemize}
    \item ResNet50 (Fig. \ref{WIKIRes})
    \begin{enumerate}
        \item $L_b = 192$. The accuracy changes within 10\% in general. However when $W_b = 6$, the accuracy decreases more significantly (over 10\%). 
        \item $L_b = 96$. The accuracy decreases with $W_b$, but beyond 65\%, with the exception at $W_b = 6$.
        \item $L_b = 48$. The accuracy falls markedly, and the accuracy does not change evidently with a decrease in width.
        \item $L_b = 24$. The accuracy is approximately 65\%, and does not vary with the decrease of $W_b$.
        \item $L_b = 12$. The accuracy is around 60\% generally, and when $W_b = 12$ the accuracy is less than 60\%.
        \item $L_b = 6$. The accuracy is similar to the situation when $L_b = 12$ that is close to 60\%, and when $W_b = 48$ the accuracy is the lowest.
    \end{enumerate}

\begin{figure}
\centering

\subfigure[$L_b$ $=$ 192]{
\begin{minipage}[t]{0.30\linewidth}
\centering
\includegraphics[width=\linewidth]{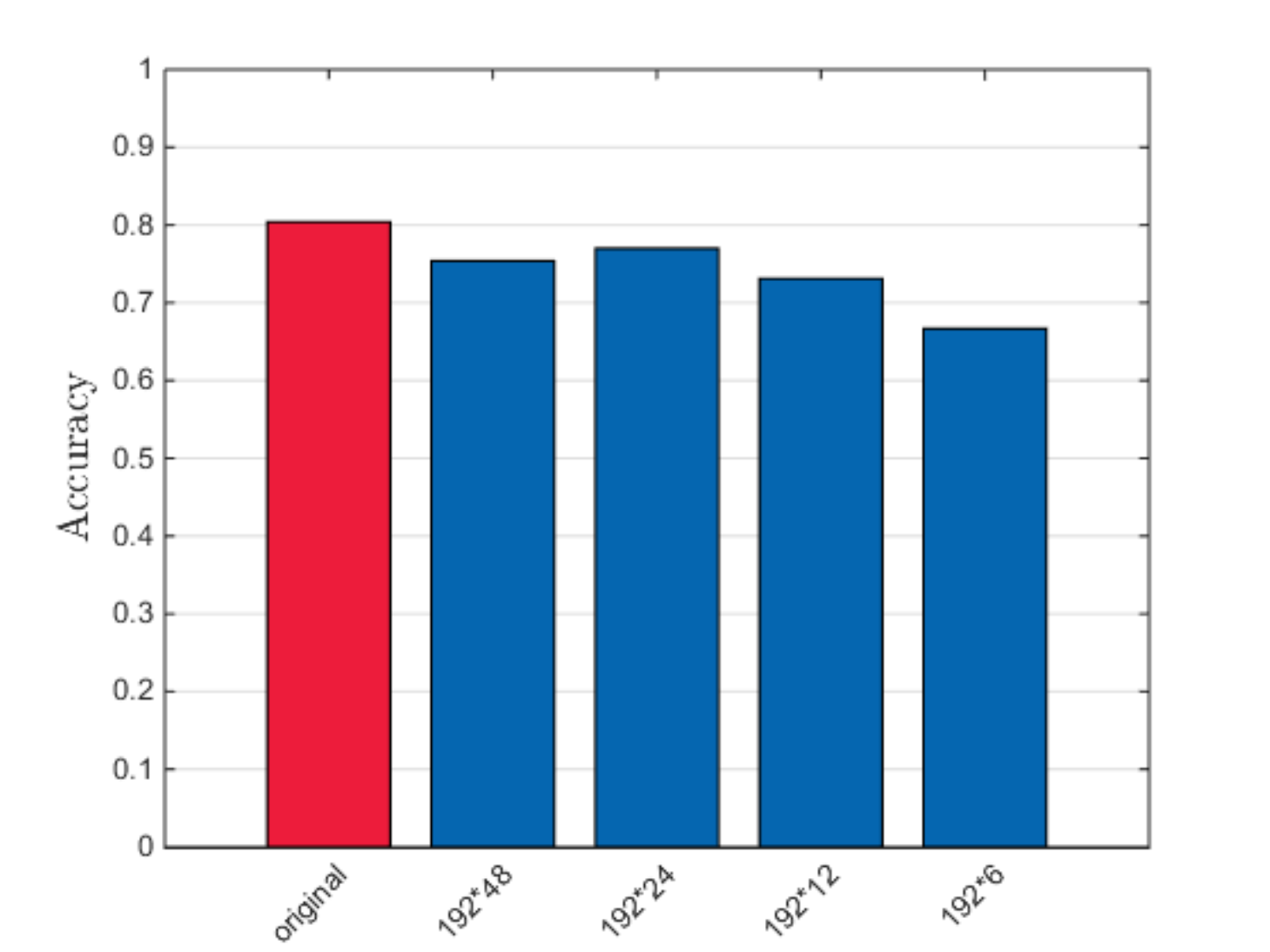}
\label{1-1}
\end{minipage}%
}%
\subfigure[$L_b$ $=$ 96]{
\begin{minipage}[t]{0.30\linewidth}
\centering
\includegraphics[width=\linewidth]{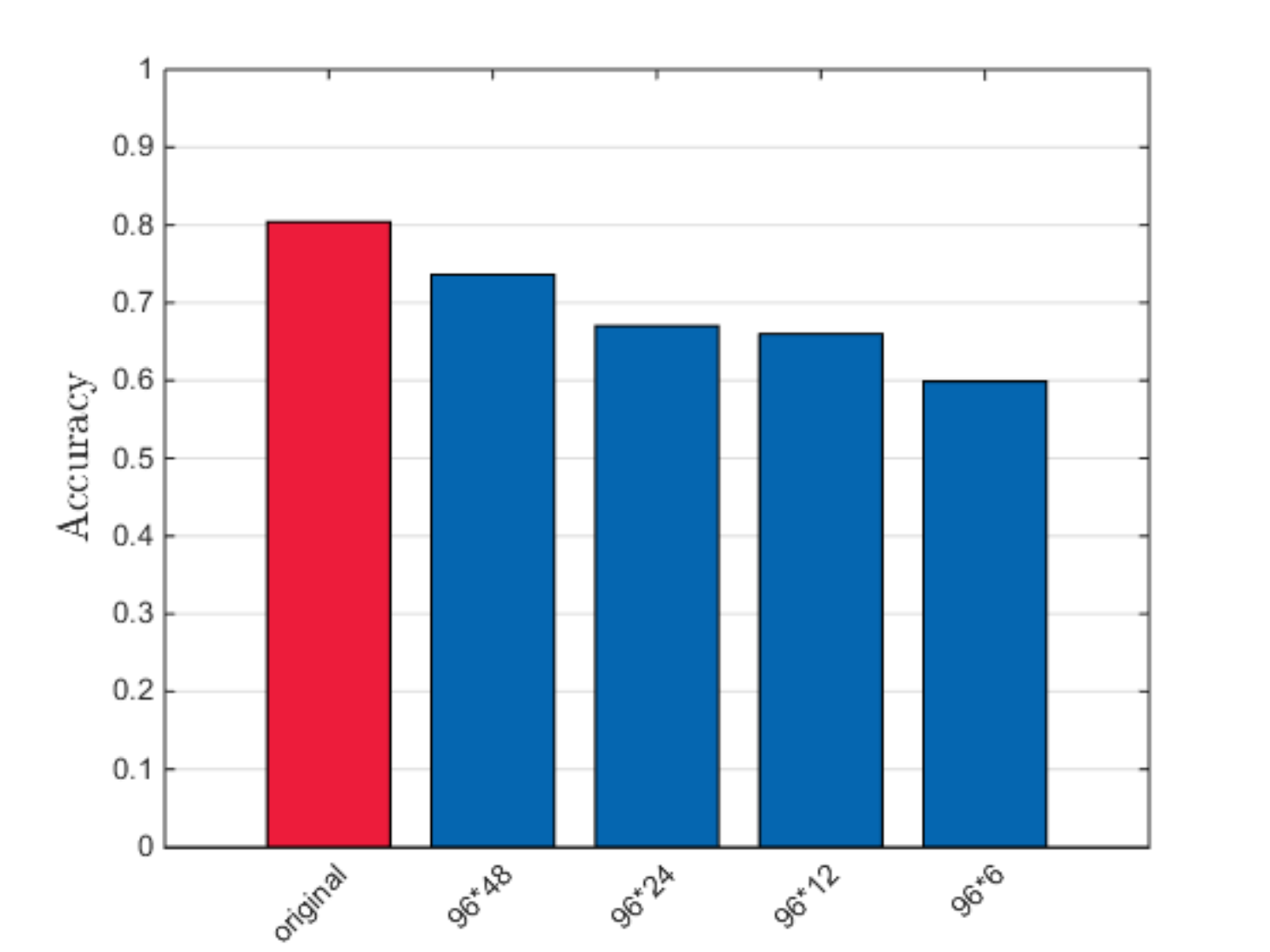}
\label{1-2}
\end{minipage}%
}%
\subfigure[$L_b$ $=$ 48]{
\begin{minipage}[t]{0.30\linewidth}
\centering
\includegraphics[width=\linewidth]{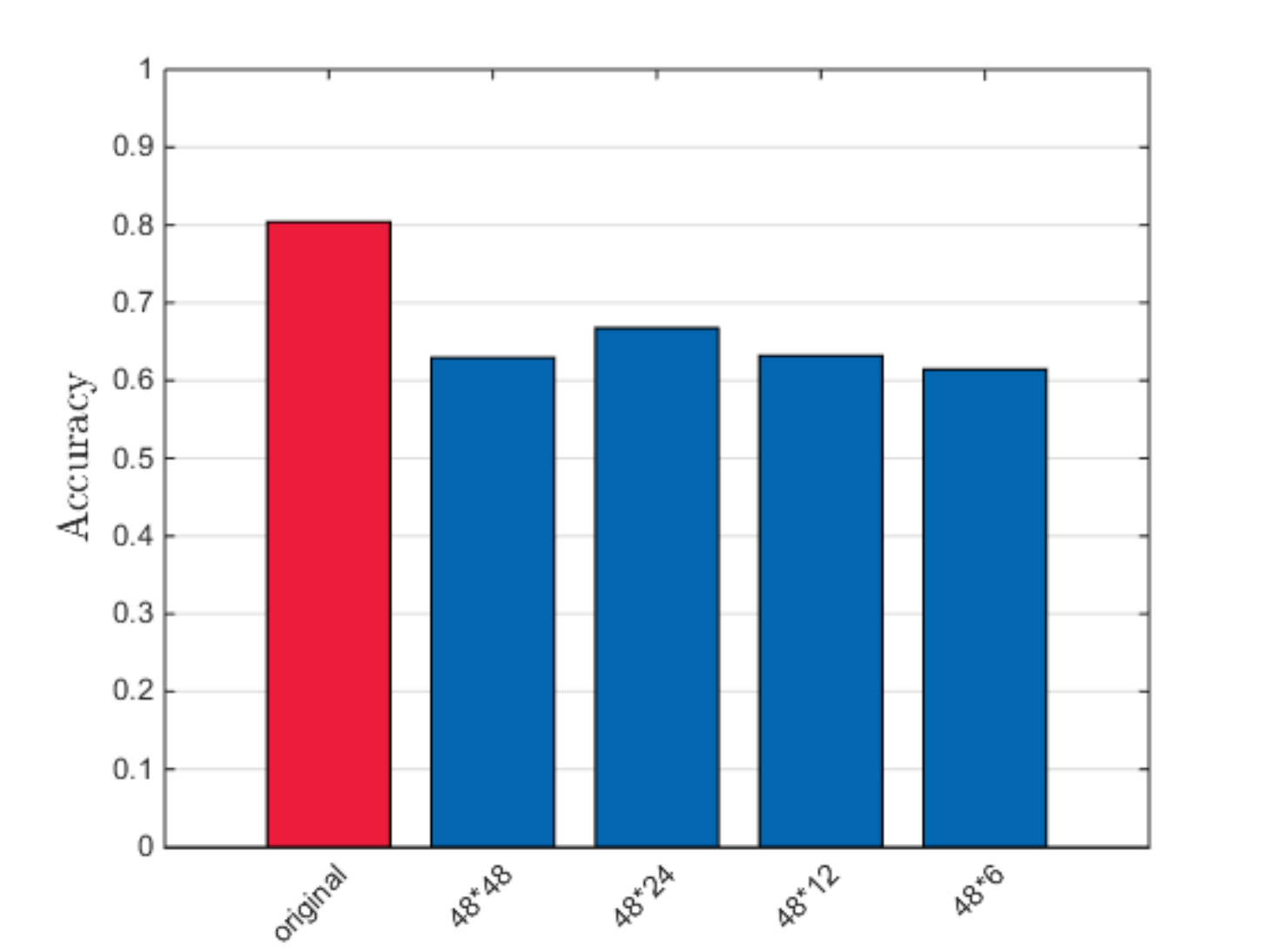}
\label{1-3}
\end{minipage}
}%
\centering
\\
\subfigure[$L_b$ $=$ 24]{
\begin{minipage}[t]{0.30\linewidth}
\centering
\includegraphics[width=\linewidth]{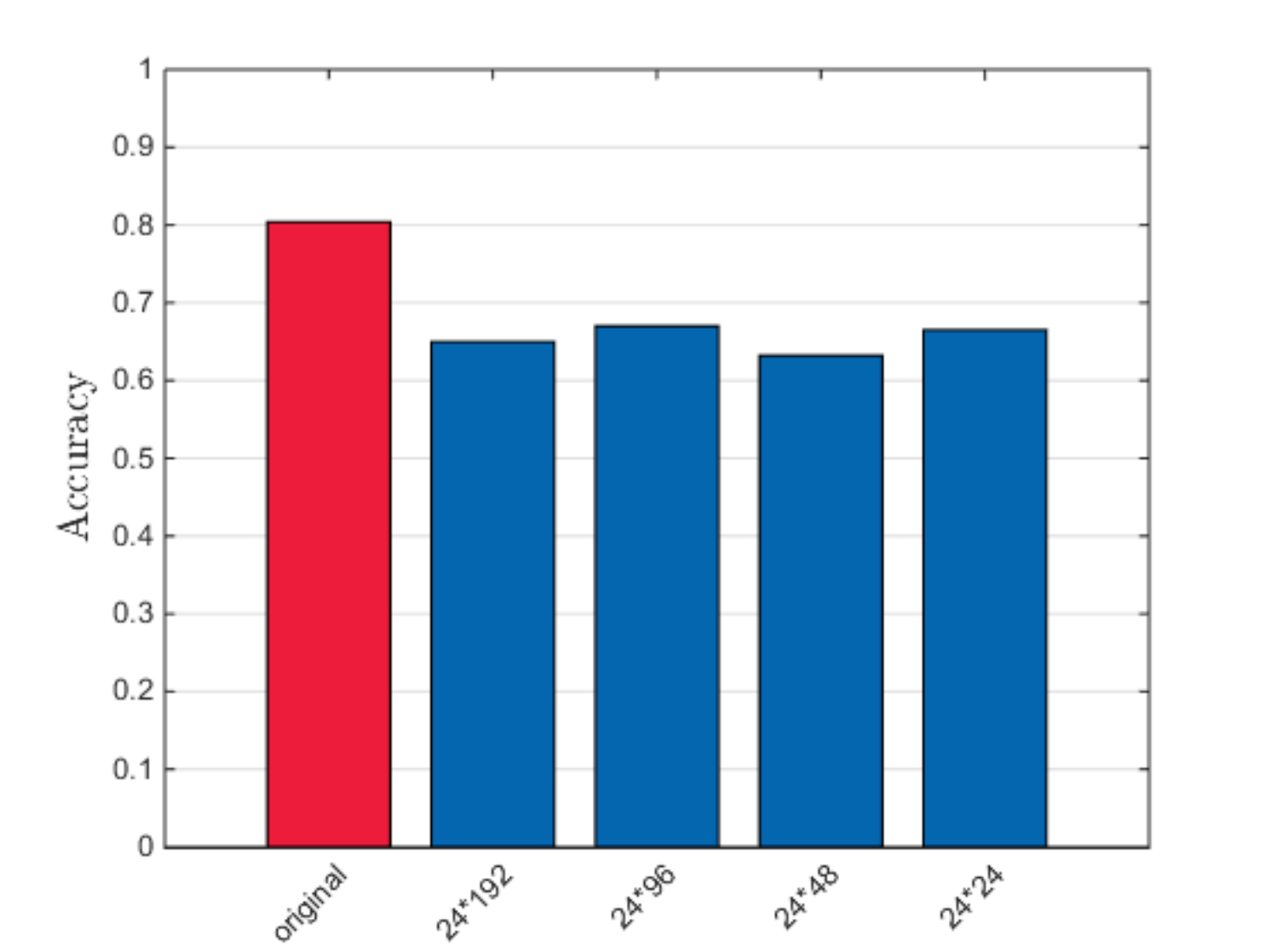}
\label{1-4}
\end{minipage}
}%
\subfigure[$L_b$ $=$ 12]{
\begin{minipage}[t]{0.30\linewidth}
\centering
\includegraphics[width=\linewidth]{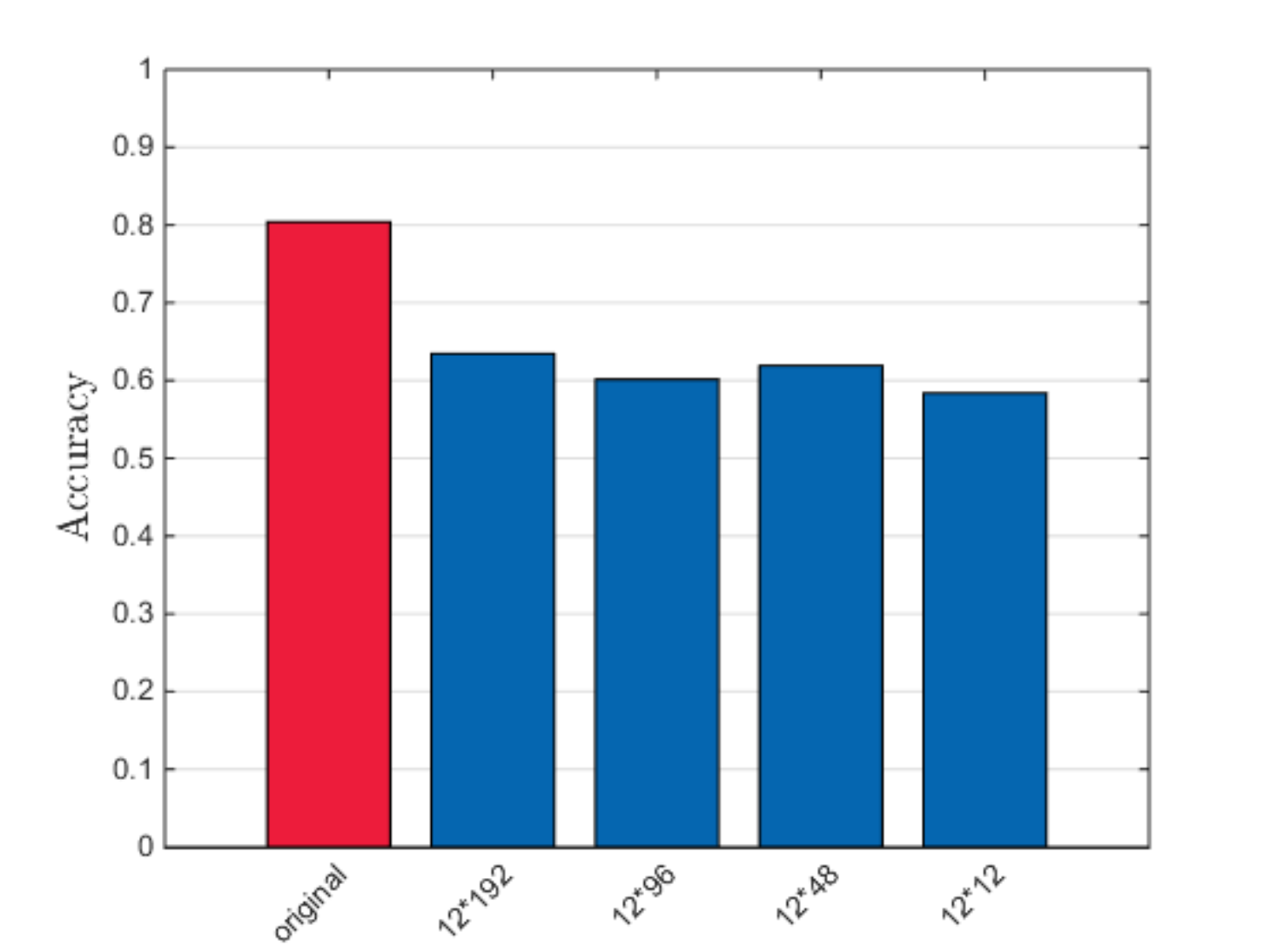}
\label{1-5}
\end{minipage}%
}%
\subfigure[$L_b$ $=$ 6]{
\begin{minipage}[t]{0.30\linewidth}
\centering
\includegraphics[width=\linewidth]{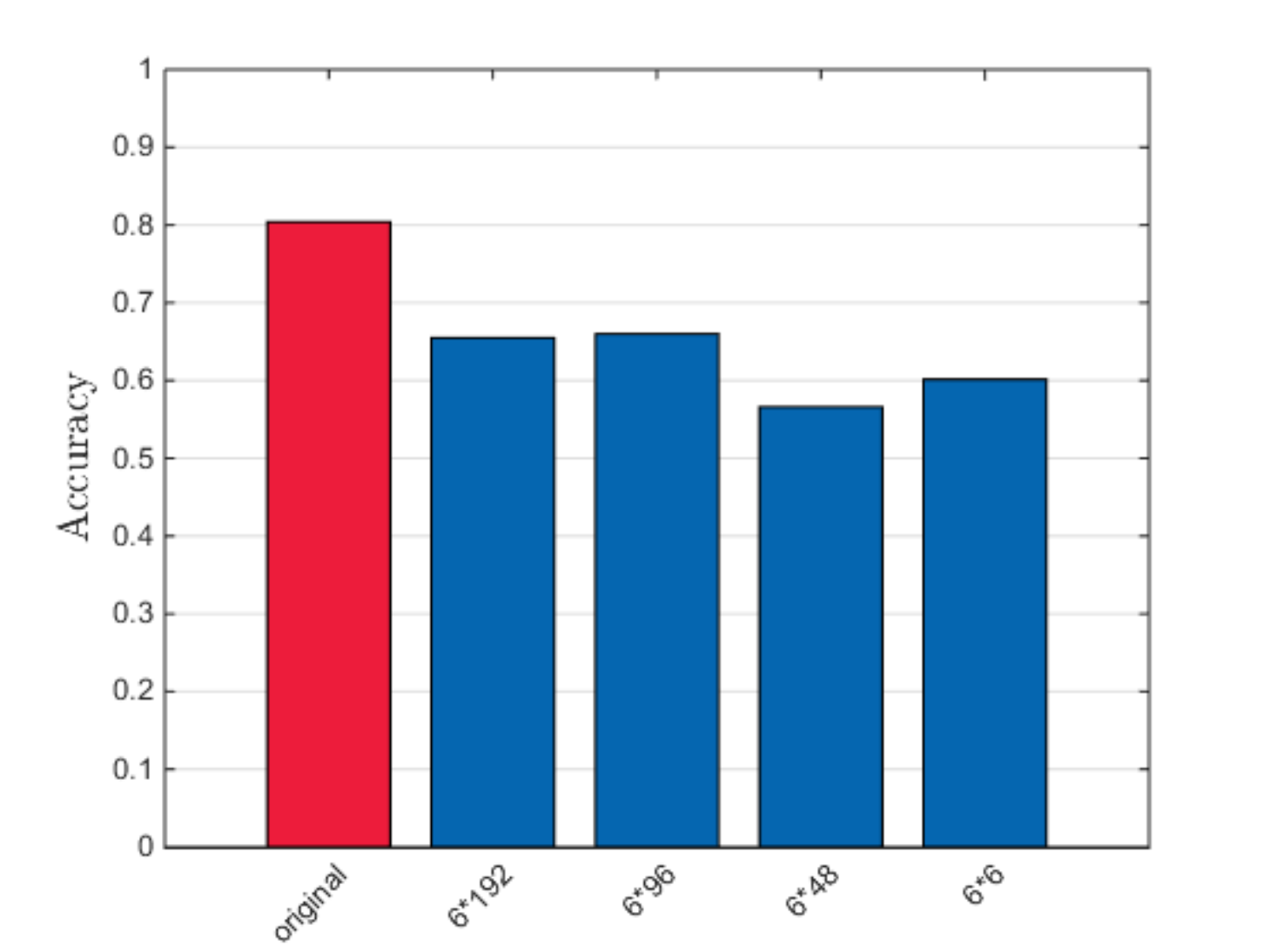}
\label{1-6}
\end{minipage}%
}%
\centering
\caption{Accuracy achieved on the WIKI dataset with different $L_b$ on ResNet50}
\label{WIKIRes}
\end{figure}
    
    
    \item VGG16 (Fig. \ref{WIKIVGG})
    \begin{enumerate}
        \item $L_b = 192$. The accuracy is essentially the same compared to the original result. When $W_b = 24$, the accuracy is even higher than the original result. 
        \item $L_b = 96$. The accuracy goes down smoothly and slowly as $W_b$ decreases, and the changes are only slight. 
        \item $L_b = 48, 24, 12, 6$. The accuracies change only approximately 5\% that are all around 70\%.
    \end{enumerate}
    
\begin{figure}
\centering

\subfigure[$L_b$ $=$ 192]{
\begin{minipage}[t]{0.30\linewidth}
\centering
\includegraphics[width=\linewidth]{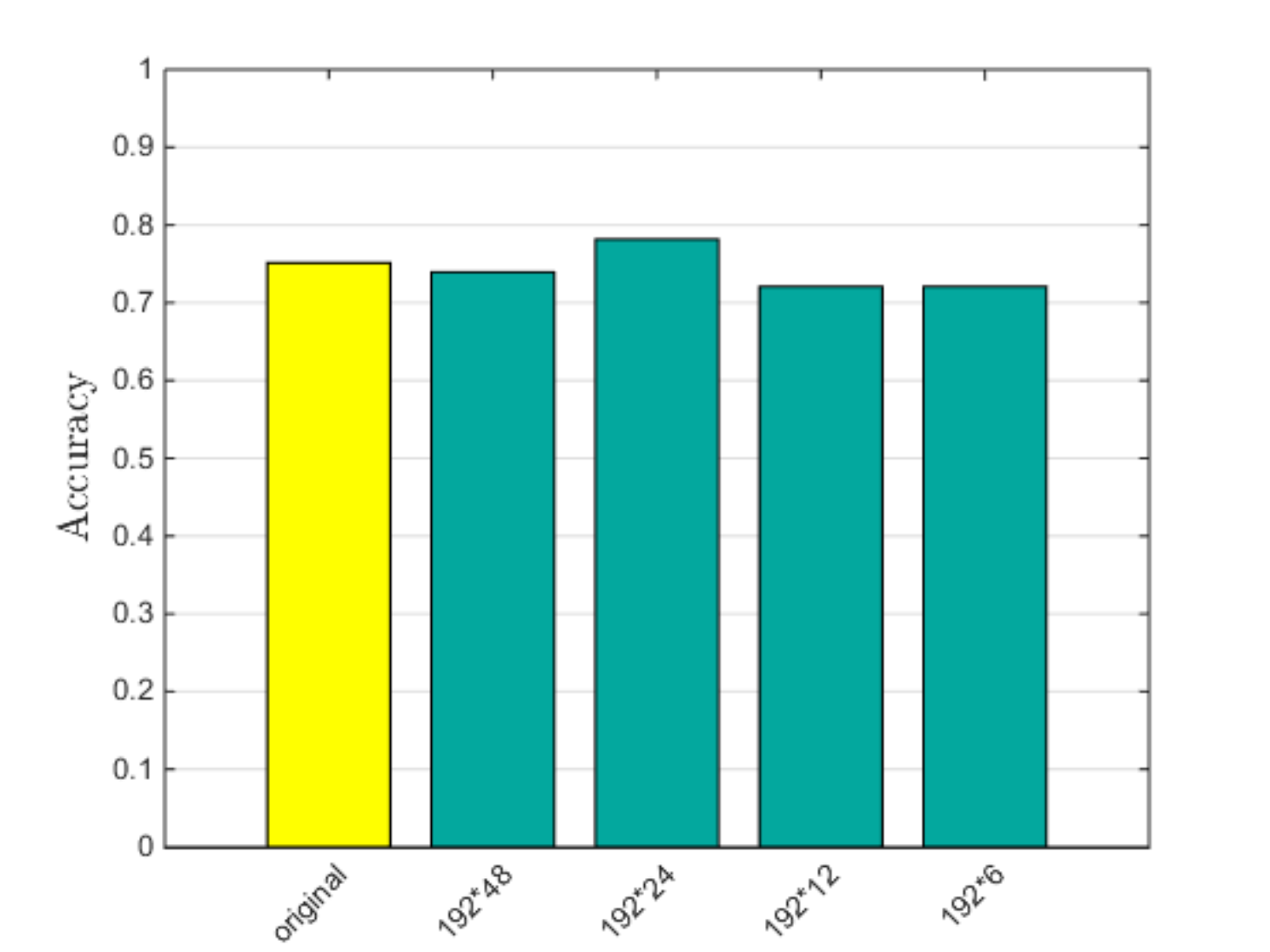}
\label{2-1}
\end{minipage}%
}%
\subfigure[$L_b$ $=$ 96]{
\begin{minipage}[t]{0.30\linewidth}
\centering
\includegraphics[width=\linewidth]{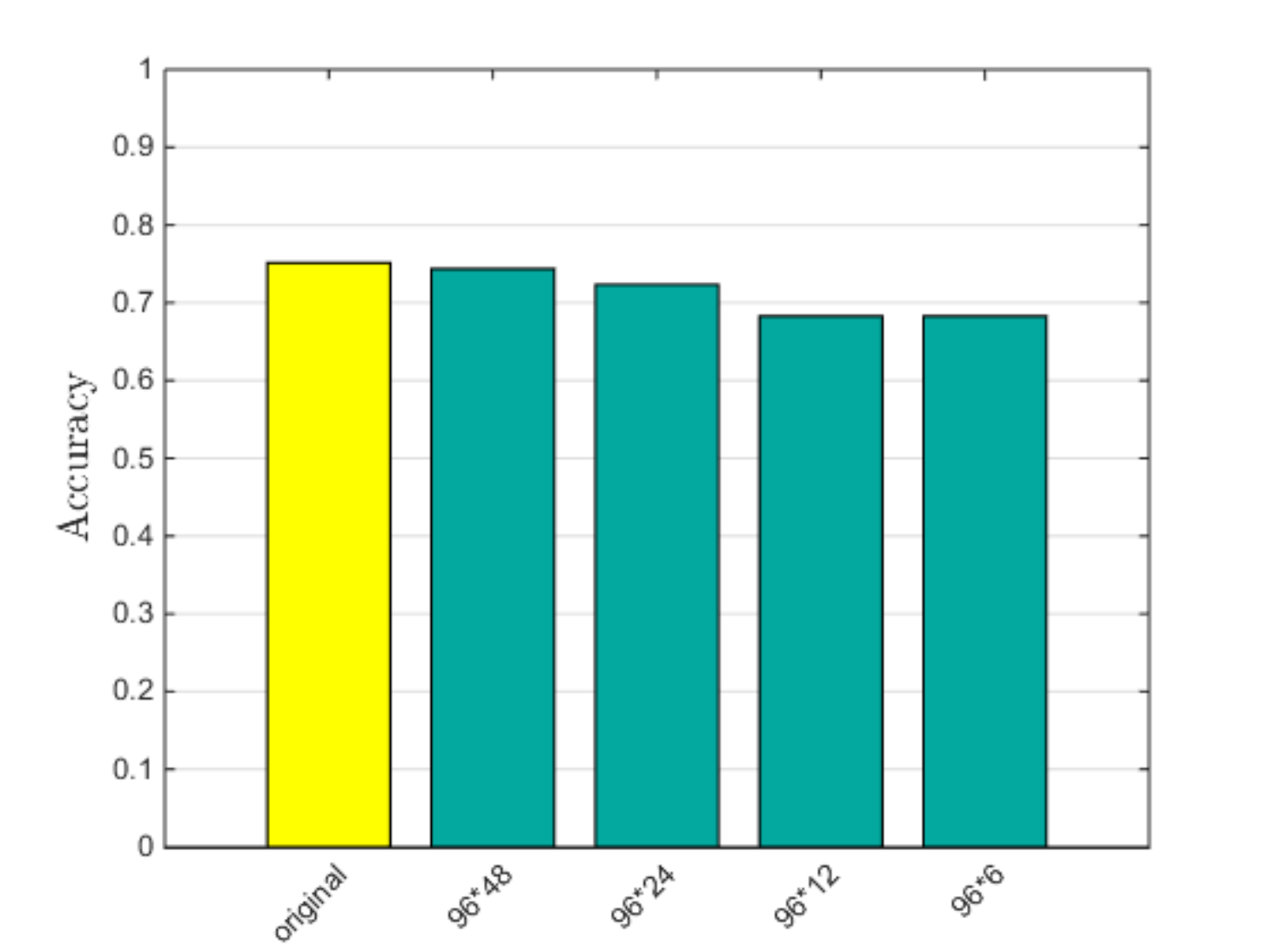}
\label{2-2}
\end{minipage}%
}%
\subfigure[$L_b$ $=$ 48]{
\begin{minipage}[t]{0.30\linewidth}
\centering
\includegraphics[width=\linewidth]{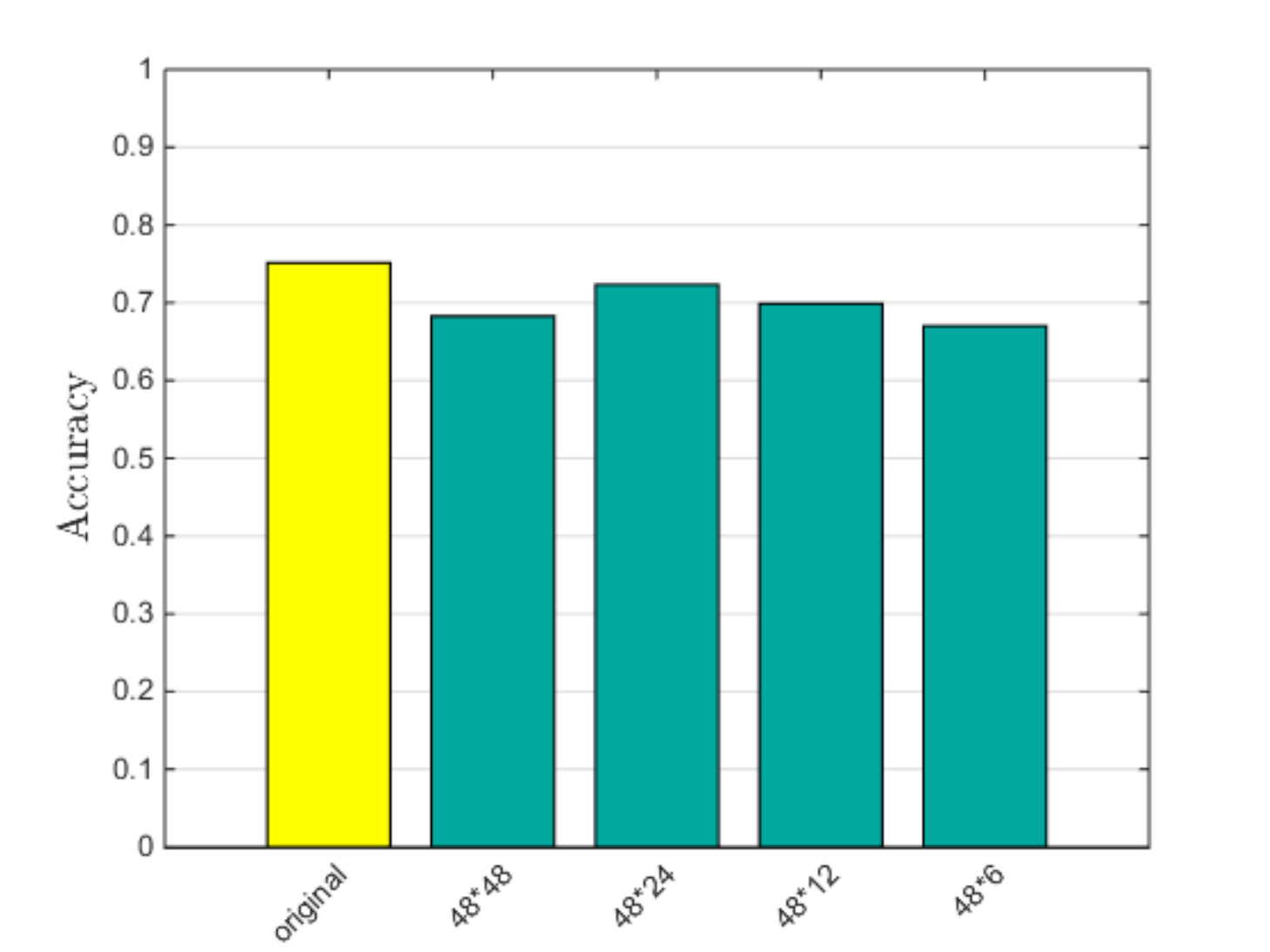}
\label{2-3}
\end{minipage}
}%
\\
\centering
\subfigure[$L_b$ $=$ 24]{
\begin{minipage}[t]{0.30\linewidth}
\centering
\includegraphics[width=\linewidth]{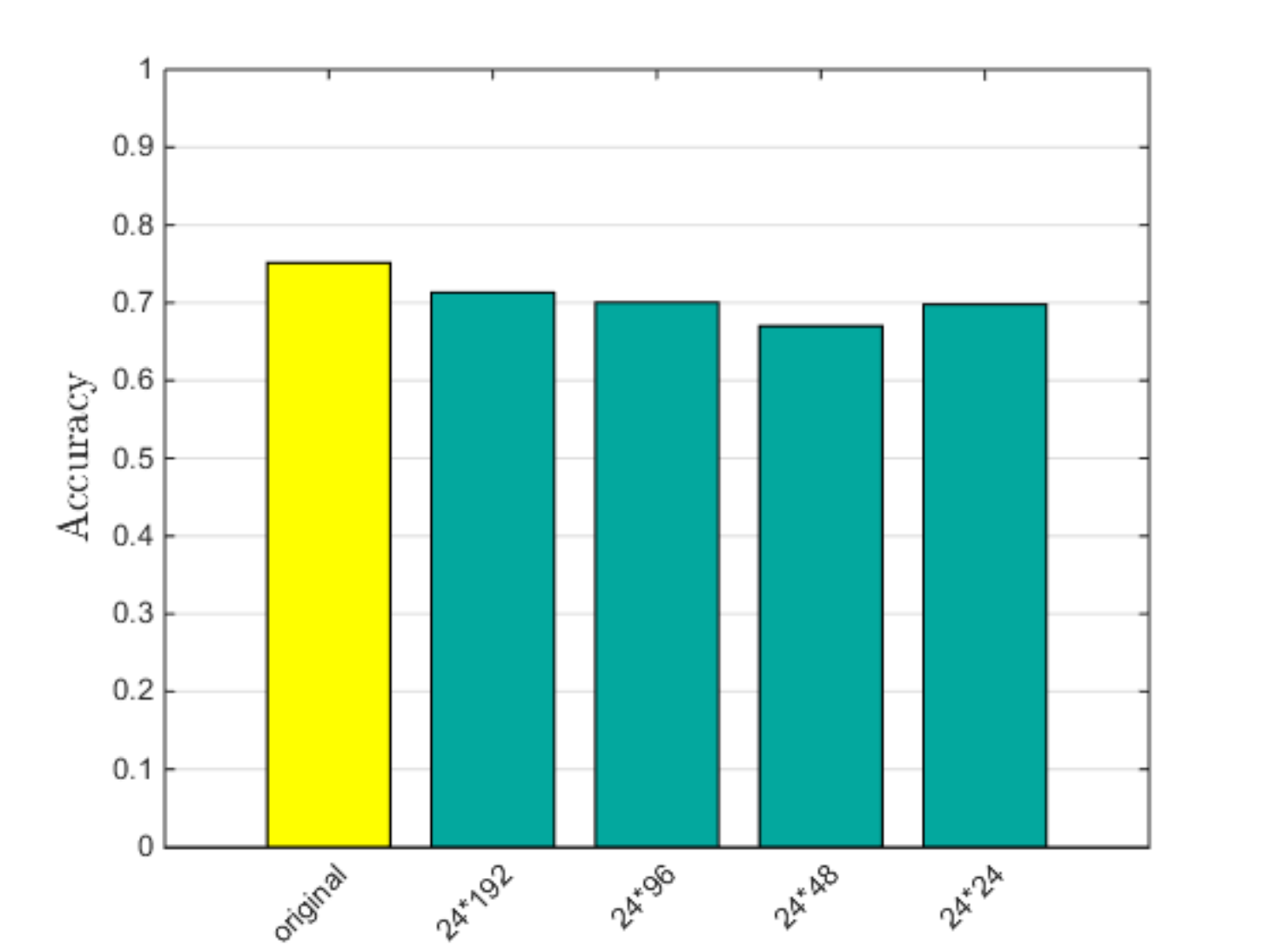}
\label{2-4}
\end{minipage}
}%
\subfigure[$L_b$ $=$ 12]{
\begin{minipage}[t]{0.30\linewidth}
\centering
\includegraphics[width=\linewidth]{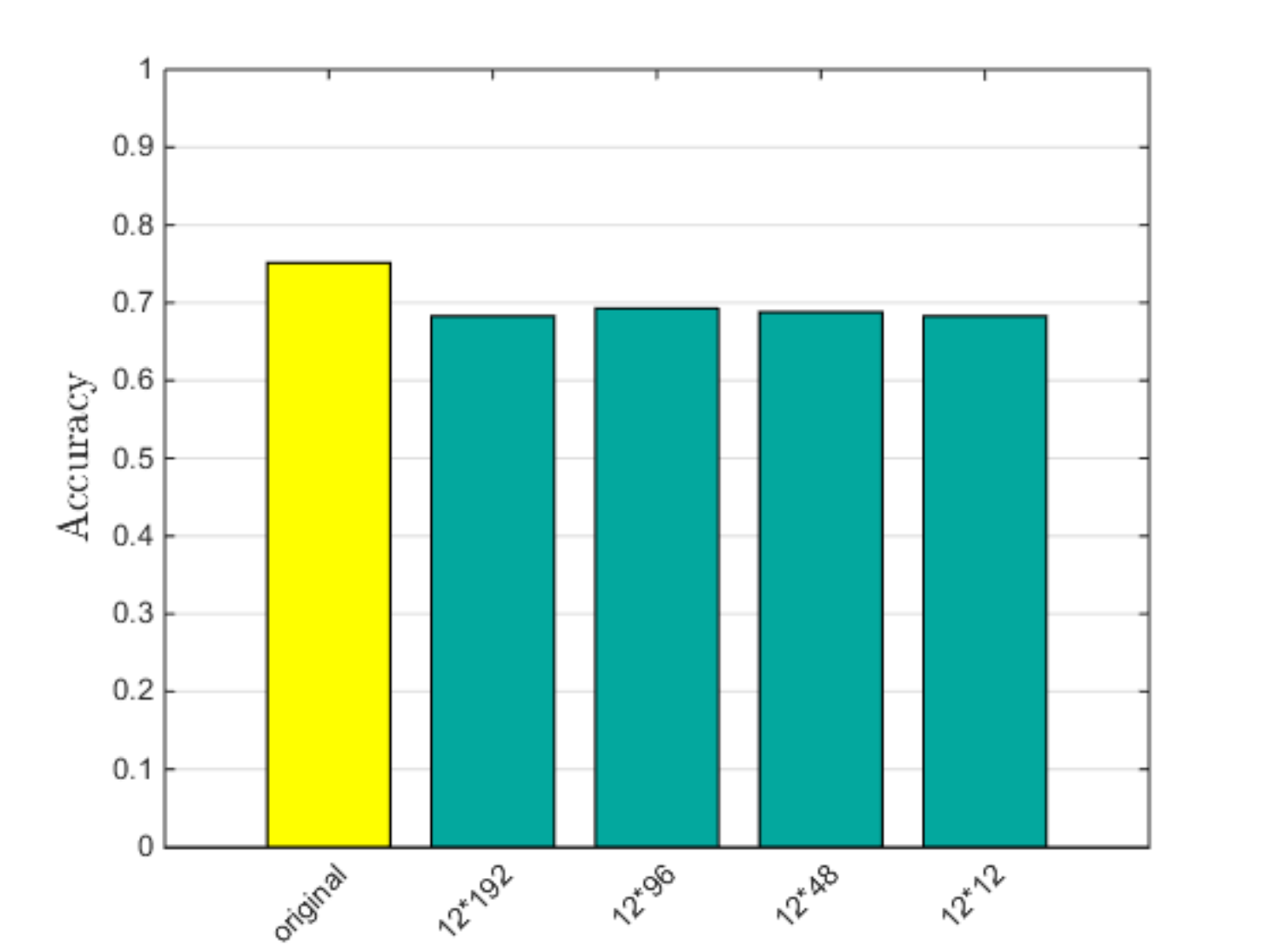}
\label{2-5}
\end{minipage}%
}%
\subfigure[$L_b$ $=$ 6]{
\begin{minipage}[t]{0.30\linewidth}
\centering
\includegraphics[width=\linewidth]{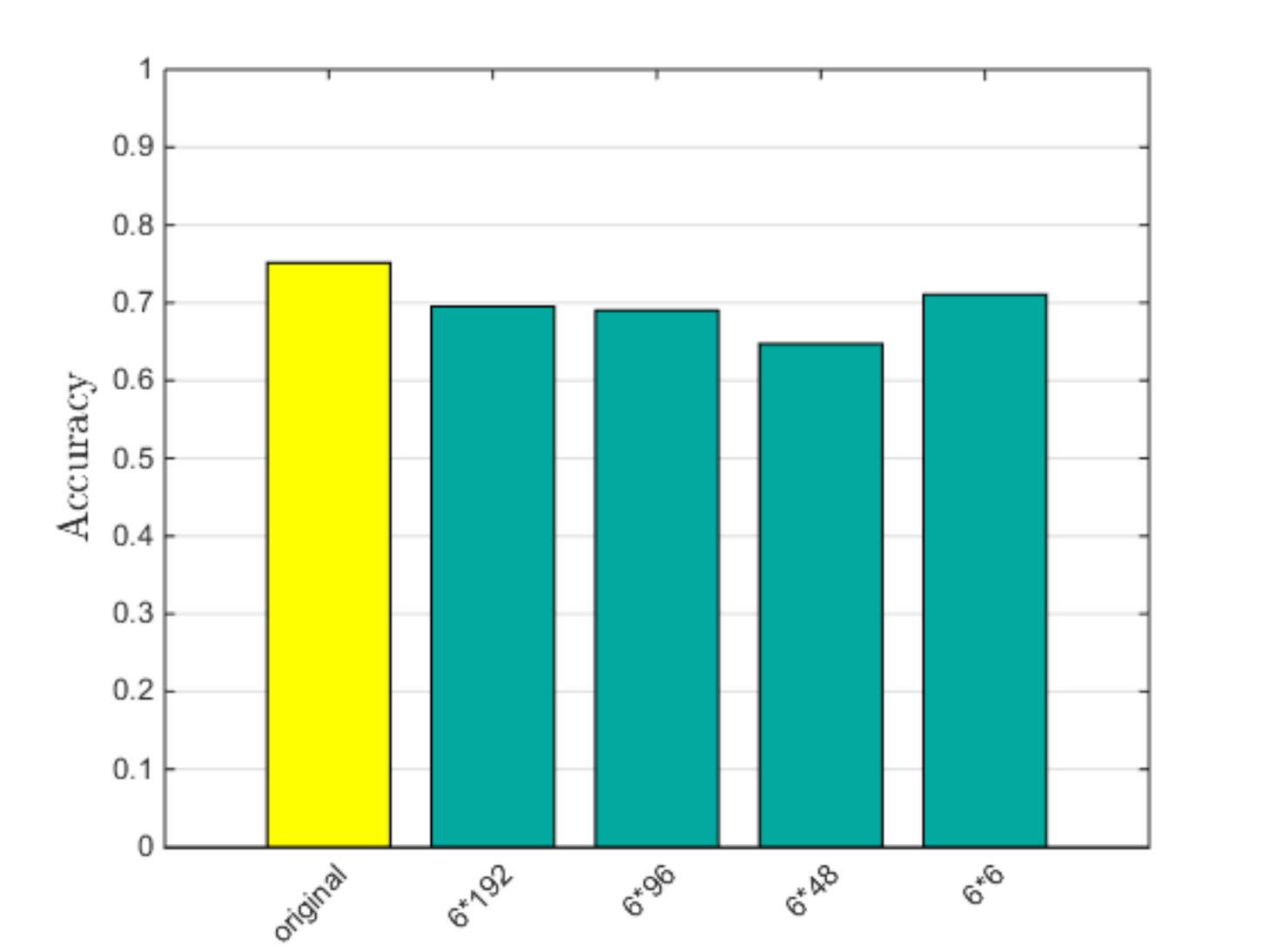}
\label{2-6}
\end{minipage}%
}%

\centering
\caption{Accuracy achieved on the WIKI dataset with different $L_b$ on VGG16}
\label{WIKIVGG}
\end{figure}    

    \item InceptionV3 (Fig. \ref{WIKIInc})
    \begin{enumerate}
        \item $L_b = 192$. The accuracy steadily declines as $W_b$ drops which remains beyond 70\%. However when $W_b = 48$, the accuracy sharply falls.
        \item $L_b = 96, 48, 24$. The accuracy decreases with $W_b$, but the accuracy is close to or beyond 70\% generally.
        \item $L_b = 12$. The accuracy decreases with $W_b$, and the accuracy is approximately 70\%, except when $W_b = 12$ (accuracy decreases significantly).
        \item $L_b = 6$. The accuracy decreases significantly compared to the previous results. When $W_b = 48, 24$, the accuracy is less than 60\%.
    \end{enumerate}
    
\begin{figure}[!htbp]
\centering

\subfigure[$L_b$ $=$ 192]{
\begin{minipage}[t]{0.30\linewidth}
\centering
\includegraphics[width=\linewidth]{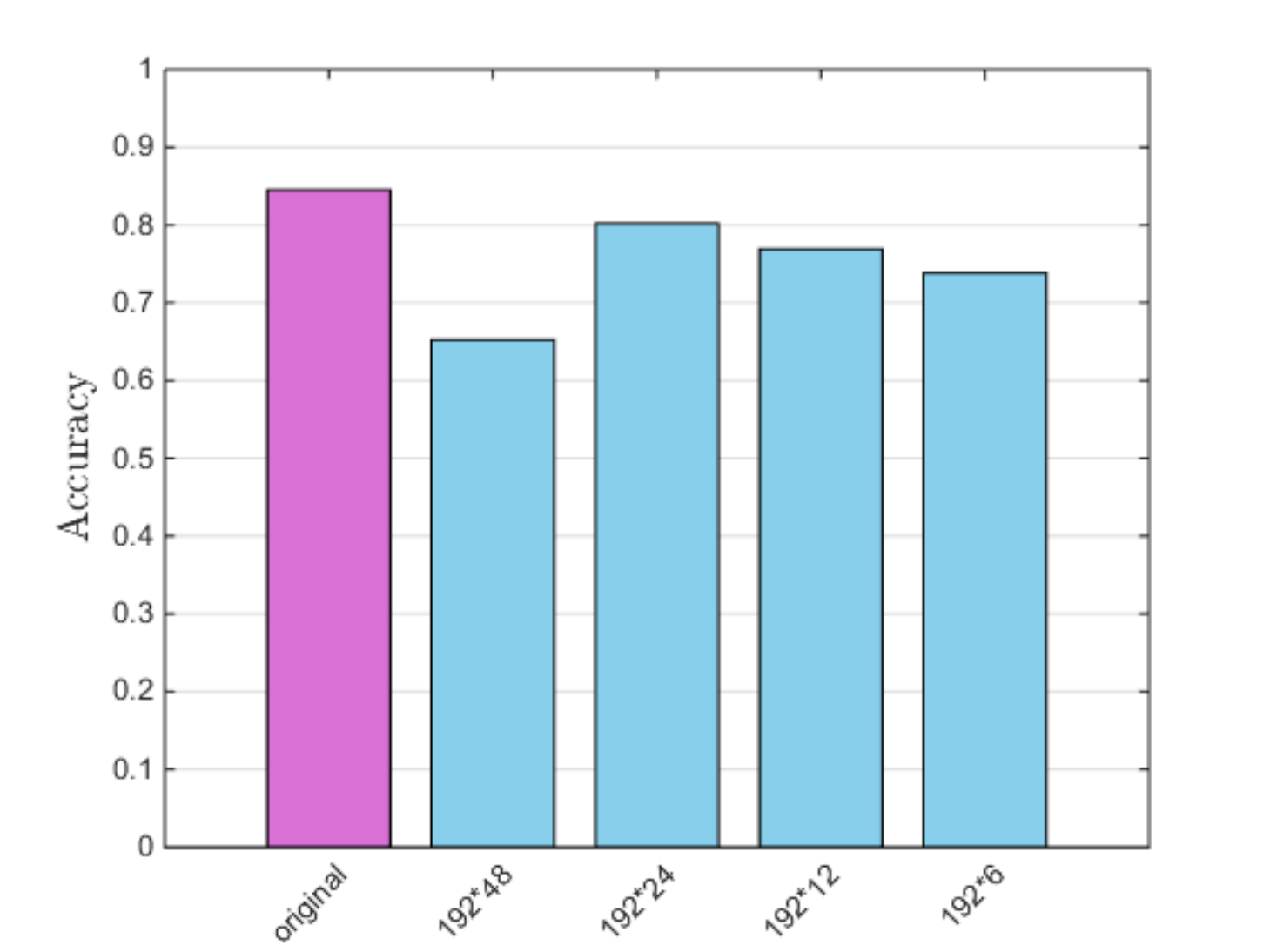}
\label{3-1}
\end{minipage}%
}%
\subfigure[$L_b$ $=$ 96]{
\begin{minipage}[t]{0.30\linewidth}
\centering
\includegraphics[width=\linewidth]{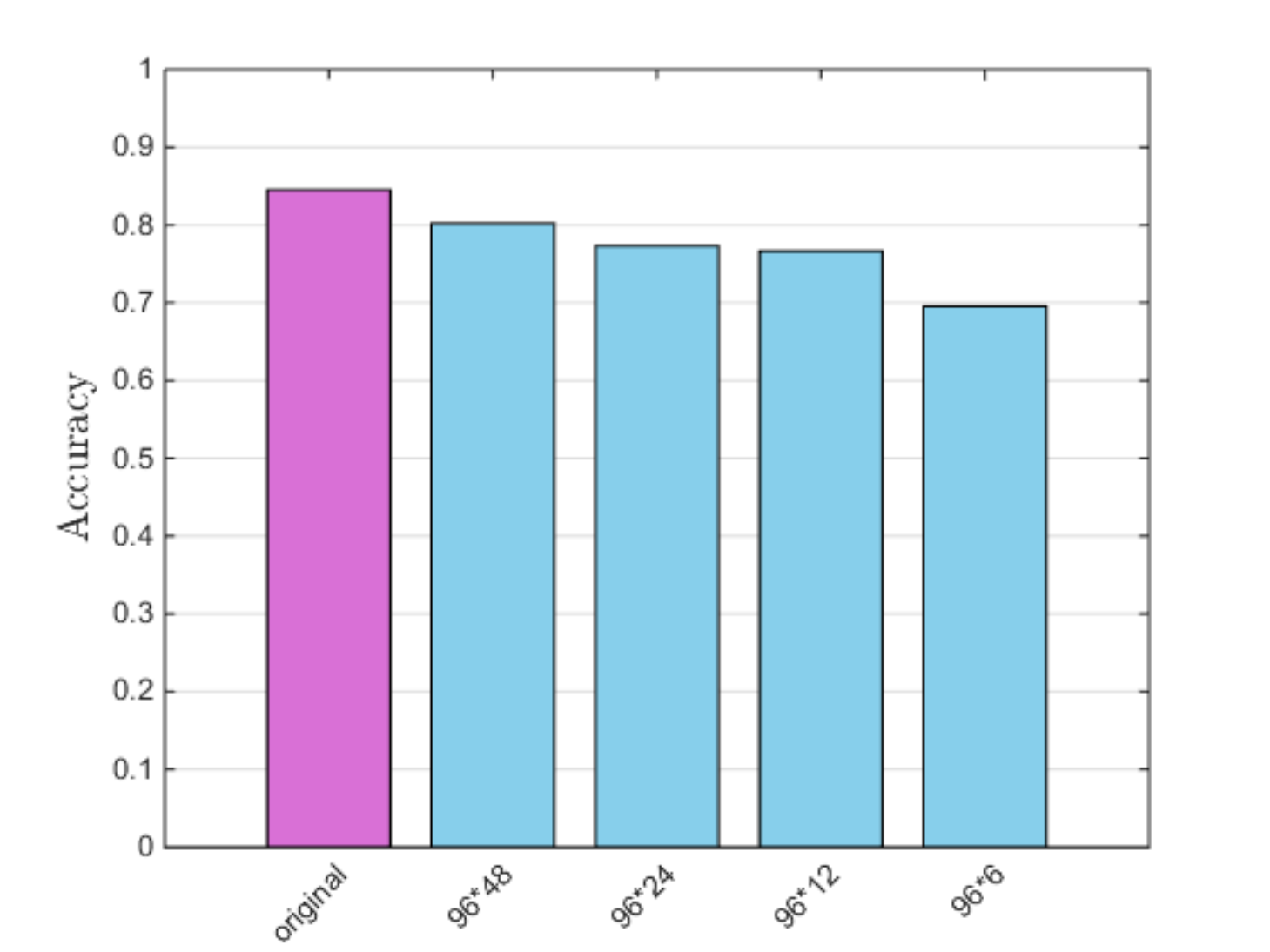}
\label{3-2}
\end{minipage}%
}%
\subfigure[$L_b$ $=$ 48]{
\begin{minipage}[t]{0.30\linewidth}
\centering
\includegraphics[width=\linewidth]{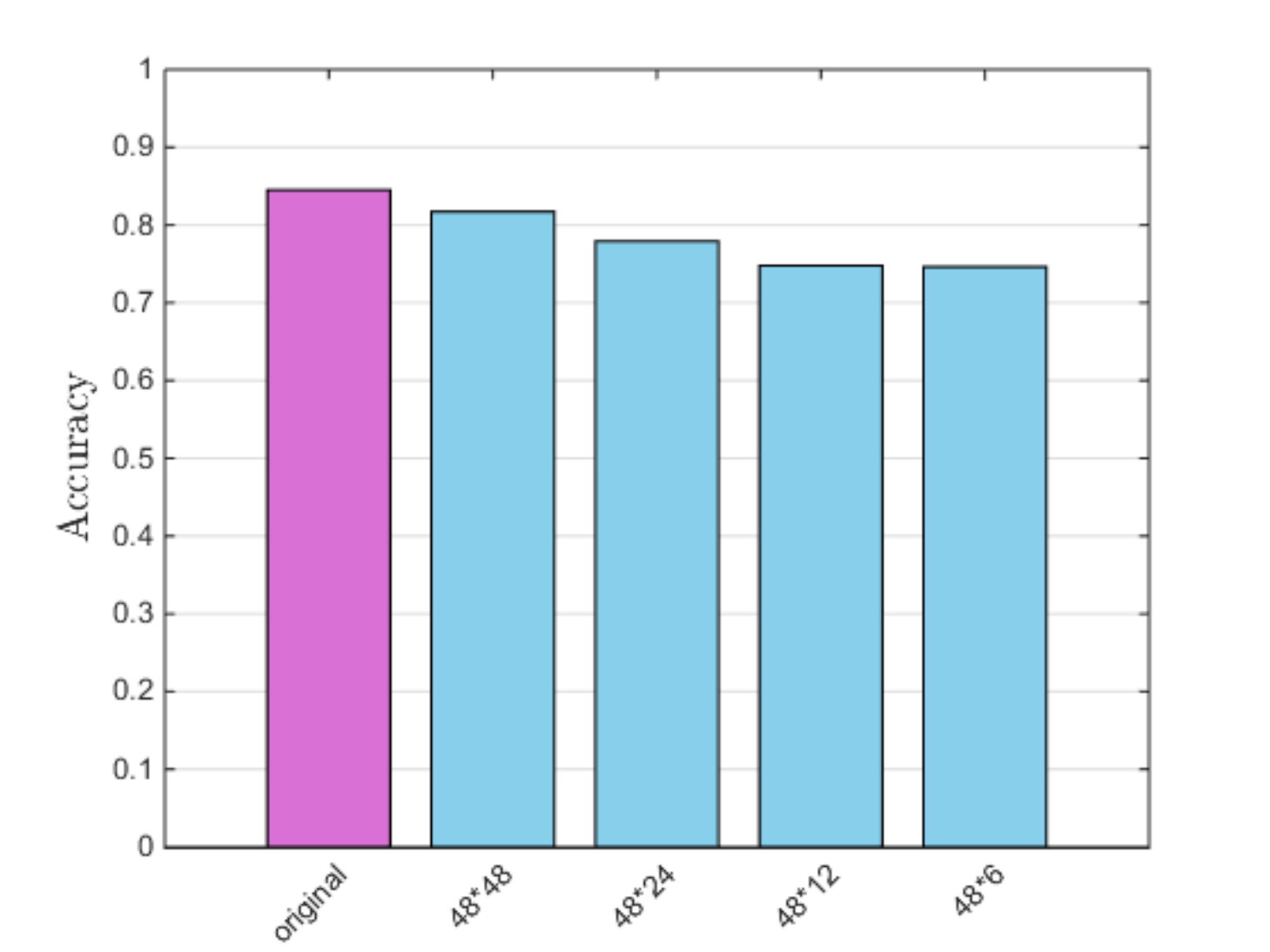}
\label{3-3}
\end{minipage}
}%
\\
\centering
\subfigure[$L_b$ $=$ 24]{
\begin{minipage}[t]{0.30\linewidth}
\centering
\includegraphics[width=\linewidth]{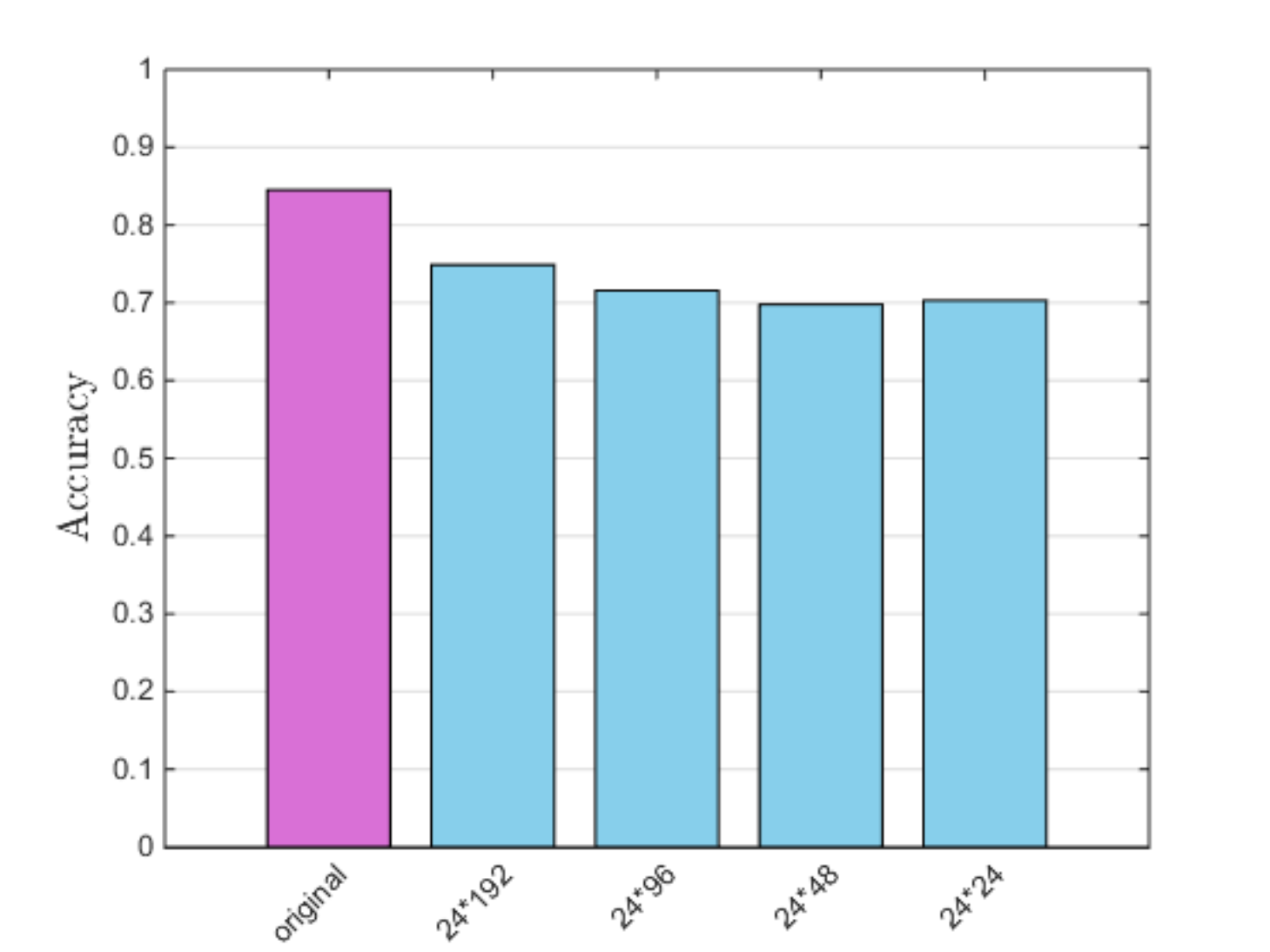}
\label{3-4}
\end{minipage}
}%
\subfigure[$L_b$ $=$ 12]{
\begin{minipage}[t]{0.30\linewidth}
\centering
\includegraphics[width=\linewidth]{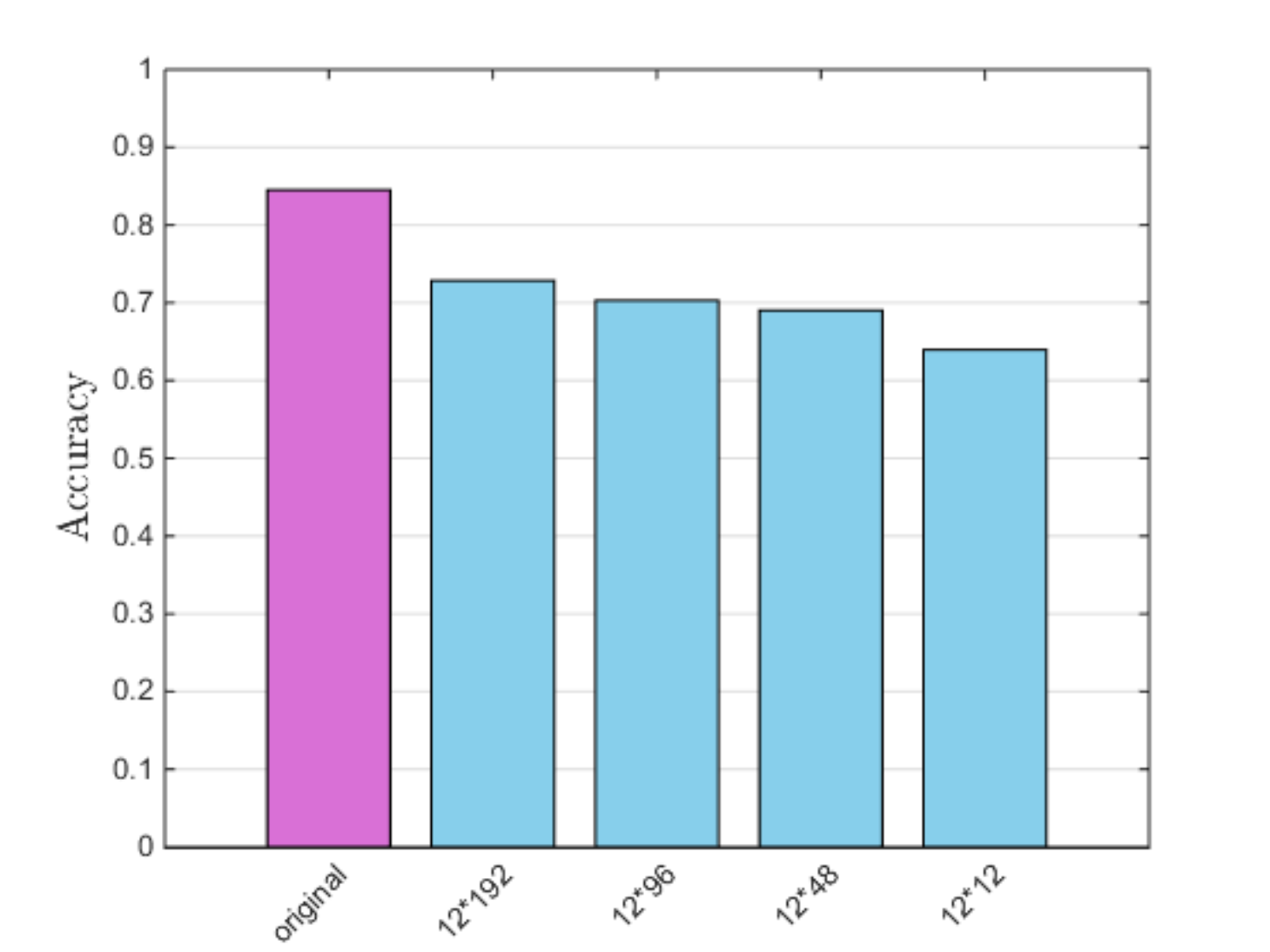}
\label{3-5}
\end{minipage}%
}%
\subfigure[$L_b$ $=$ 6]{
\begin{minipage}[t]{0.30\linewidth}
\centering
\includegraphics[width=\linewidth]{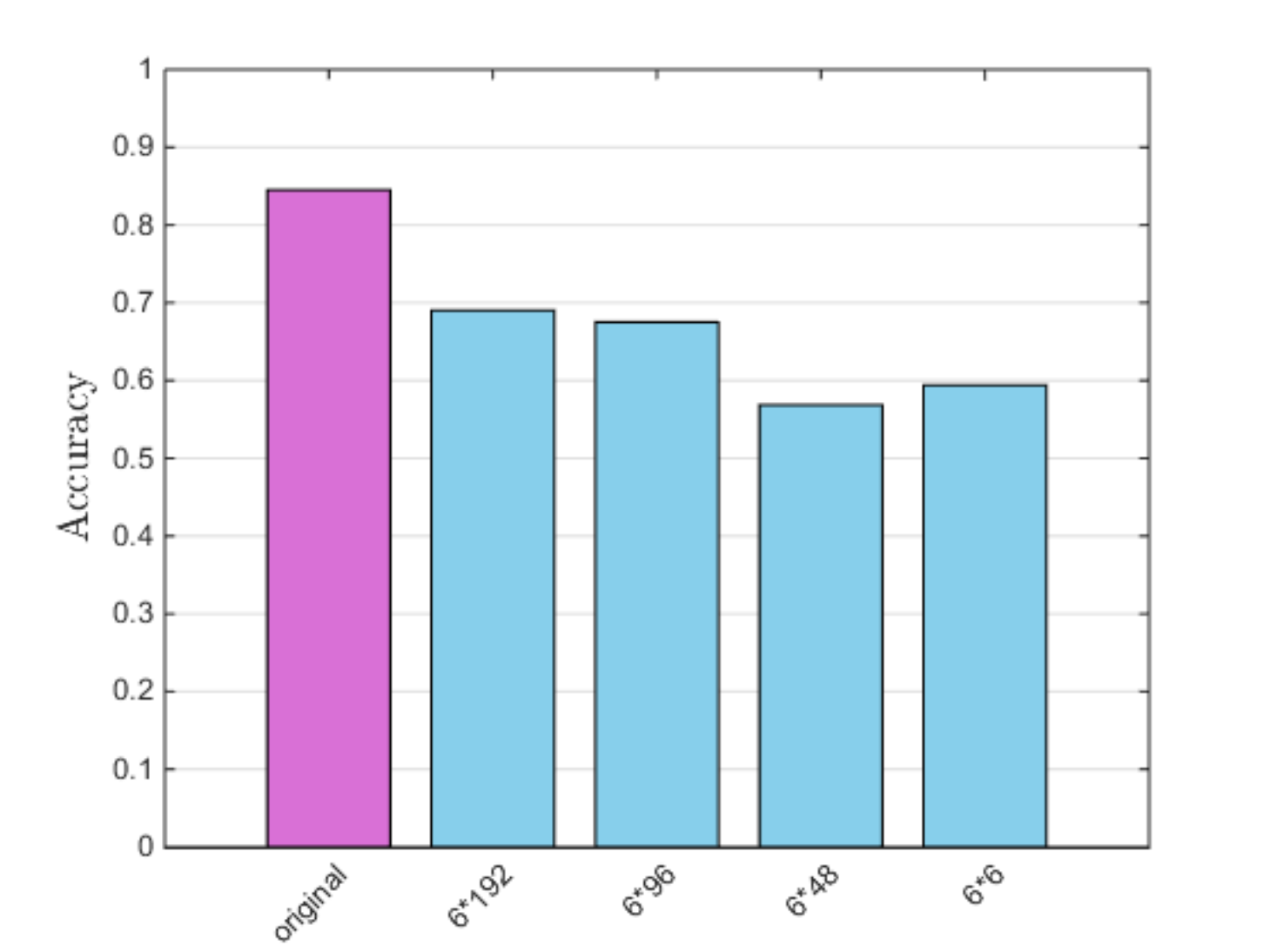}
\label{3-6}
\end{minipage}%
}%

\centering
\caption{Accuracy achieved on the WIKI dataset with different $L_b$ on InceptionV3}
\label{WIKIInc}
\end{figure}
\end{itemize}

In summary, for the WIKI dataset, the accuracy on the testing set of ResNet50 and InceptionV3 models trained by the mixed training set decreases significantly as $L_b$ and $W_b$ declines, while the loss is greater too. On the contrary, for the VGG16 model, the accuracy on the testing set is quite closer to the original results while the size of the divided pixel blocks has little effect on accuracy. 

Moreover, on ResNet50 and InceptionV3 models, when $N_b$ is determined, we calculate the average of data with the same number of pixel blocks, and we compare each case with the average. We find that when the value of $L_b$ or $W_b$ is quite small, the accuracy is lower than the average. Thus, a particularly large and unstable loss is generated, when the division parameters are small. 

Generally, on the WIKI dataset our scheme can keep an acceptable data availability on ResNet50 and InceptionV3 models for several division parameters, and higher data availability on VGG16 model for most division parameters.

\subsection{CNBC Face Dataset Experimental Results}

We apply the CNBC face dataset as a multiple classification dataset for the experiment, and we use race as the classification criterion. We test ResNet50 and DenseNet121 on the CNBC dataset. Additionally, we record and analyze the accuracy of ResNet50 and DenseNet121 models on the testing set by means of the same method in the WIKI dataset.

\begin{figure}[!htbp]
  \centering
  \includegraphics[width=0.9\linewidth]{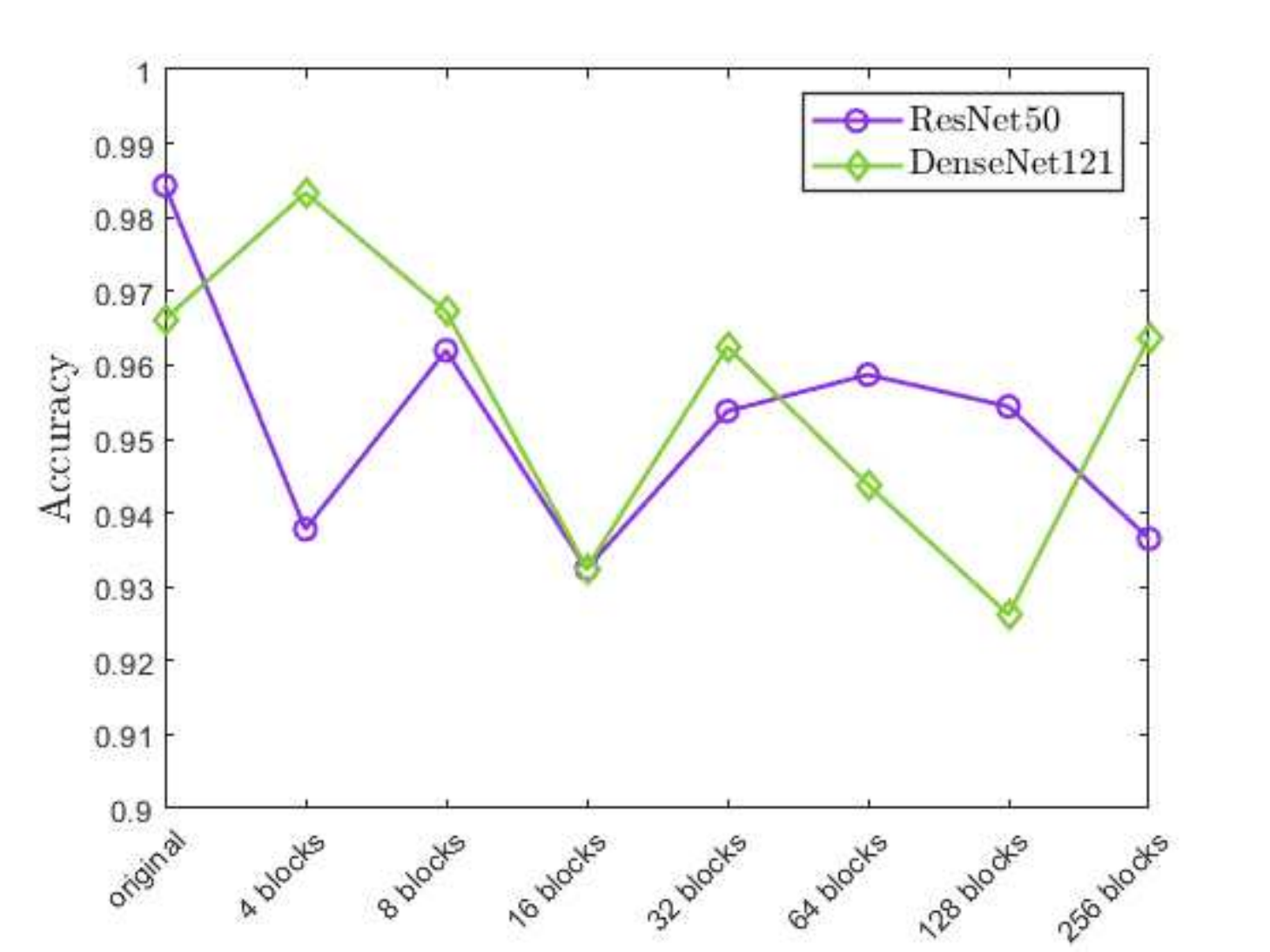}
  \caption{Trend of accuracy on the CNBC dataset}
  \label{CNBCtrand}
\end{figure}

Fig. \ref{CNBCtrand} illustrates that the accuracy of both models trained by the mixed training set on the testing set is only slightly lower than the original results (within 5\%), and the accuracies are all beyond 92\%. Besides, the loss of ResNet50 and DenseNet121 models are extremely small (fewer than $1e-5$), thus, we do not display it through the diagram.

In addition, we discuss the specific accuracy results of ResNet50 and DenseNet121 that are trained by $I_T$ with different $L_b$ and $W_b$ on the testing set. We also group the results above according to $L_b$ and compare them to the original results. We will illustrate and analyze the experimental results below:

\begin{itemize}
    \item ResNet50 (Fig.  \ref{CNBCRes})
    \begin{enumerate}
        \item $L_b = 96$. The accuracy is beyond 93\% and accuracy changes are within 5\%. In addition, accuracy drops obviously when $W_b = 24$ compared to the situations when $W_b = 12, 6$. 
        \item $L_b = 48, 24$. The accuracies held steady around 95\%. The variations of $W_b$ only have quite a small effect on the accuracy.
    \end{enumerate}
    
\begin{figure}[!htbp]
\centering

\subfigure[$L_b$ $=$ 96]{
\begin{minipage}[t]{0.30\linewidth}
\centering
\includegraphics[width=\linewidth]{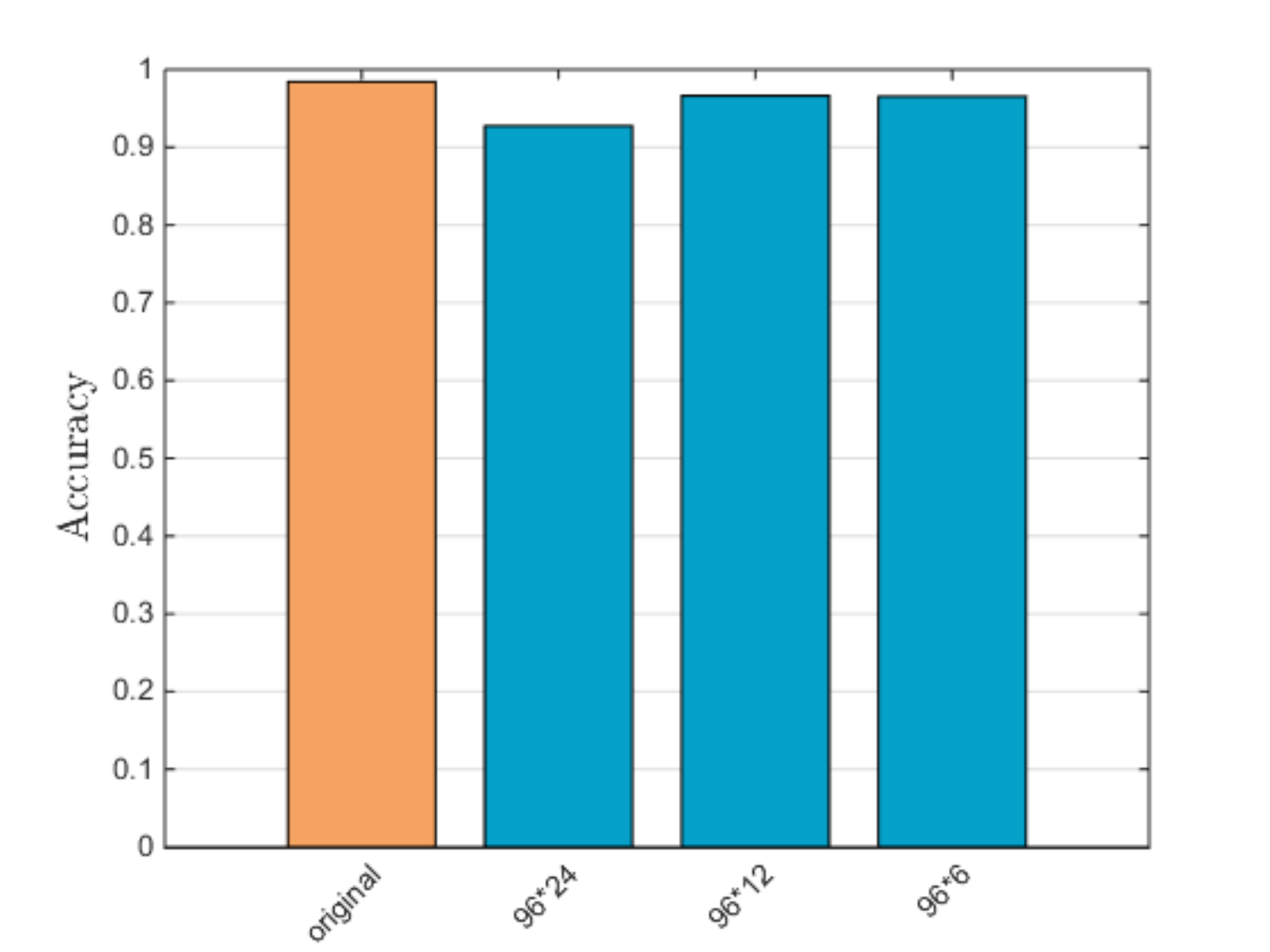}
\label{4-1}
\end{minipage}%
}%
\subfigure[$L_b$ $=$ 48]{
\begin{minipage}[t]{0.30\linewidth}
\centering
\includegraphics[width=\linewidth]{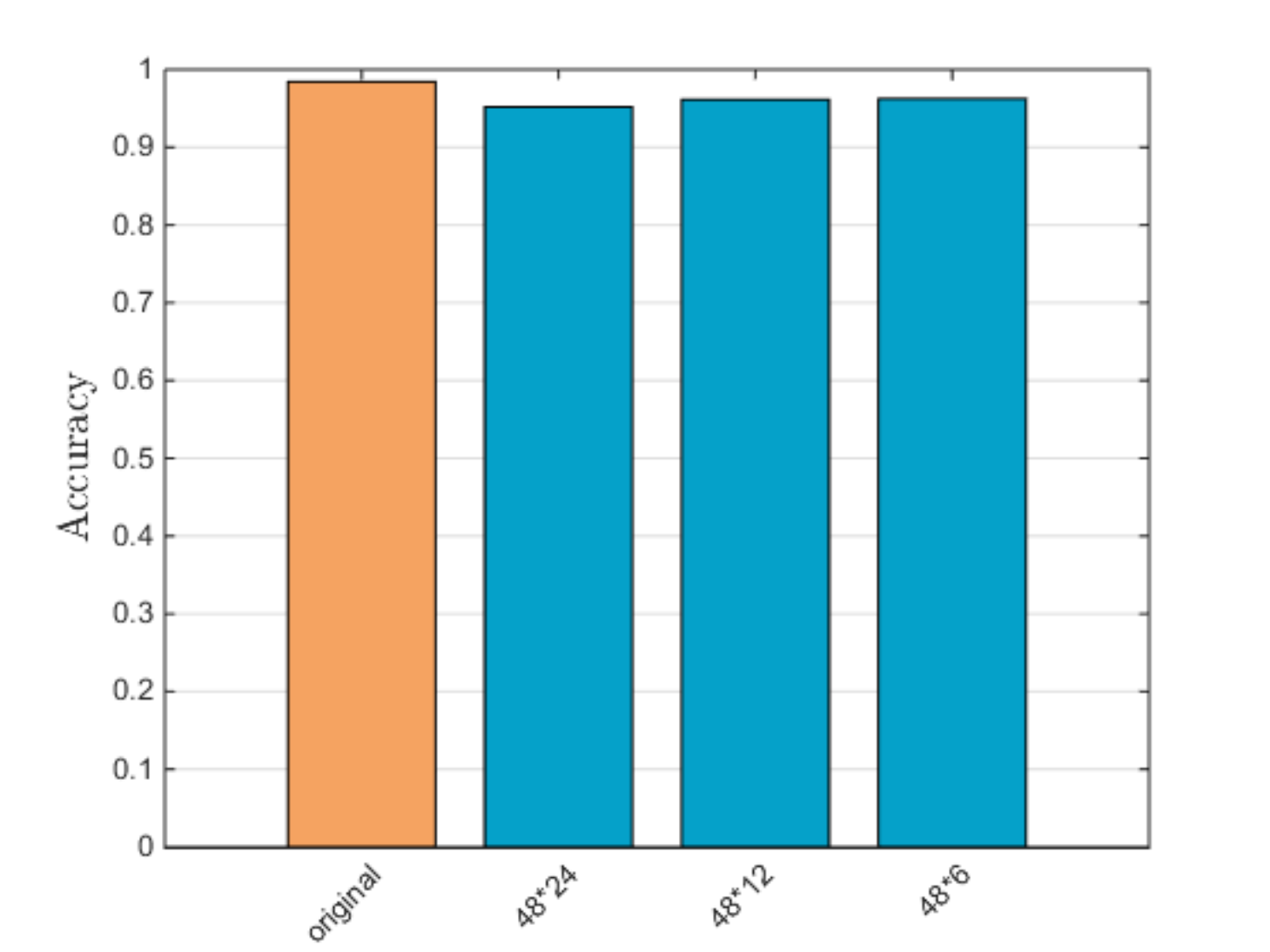}
\label{4-2}
\end{minipage}%
}%
\subfigure[$L_b$ $=$ 24]{
\begin{minipage}[t]{0.30\linewidth}
\centering
\includegraphics[width=\linewidth]{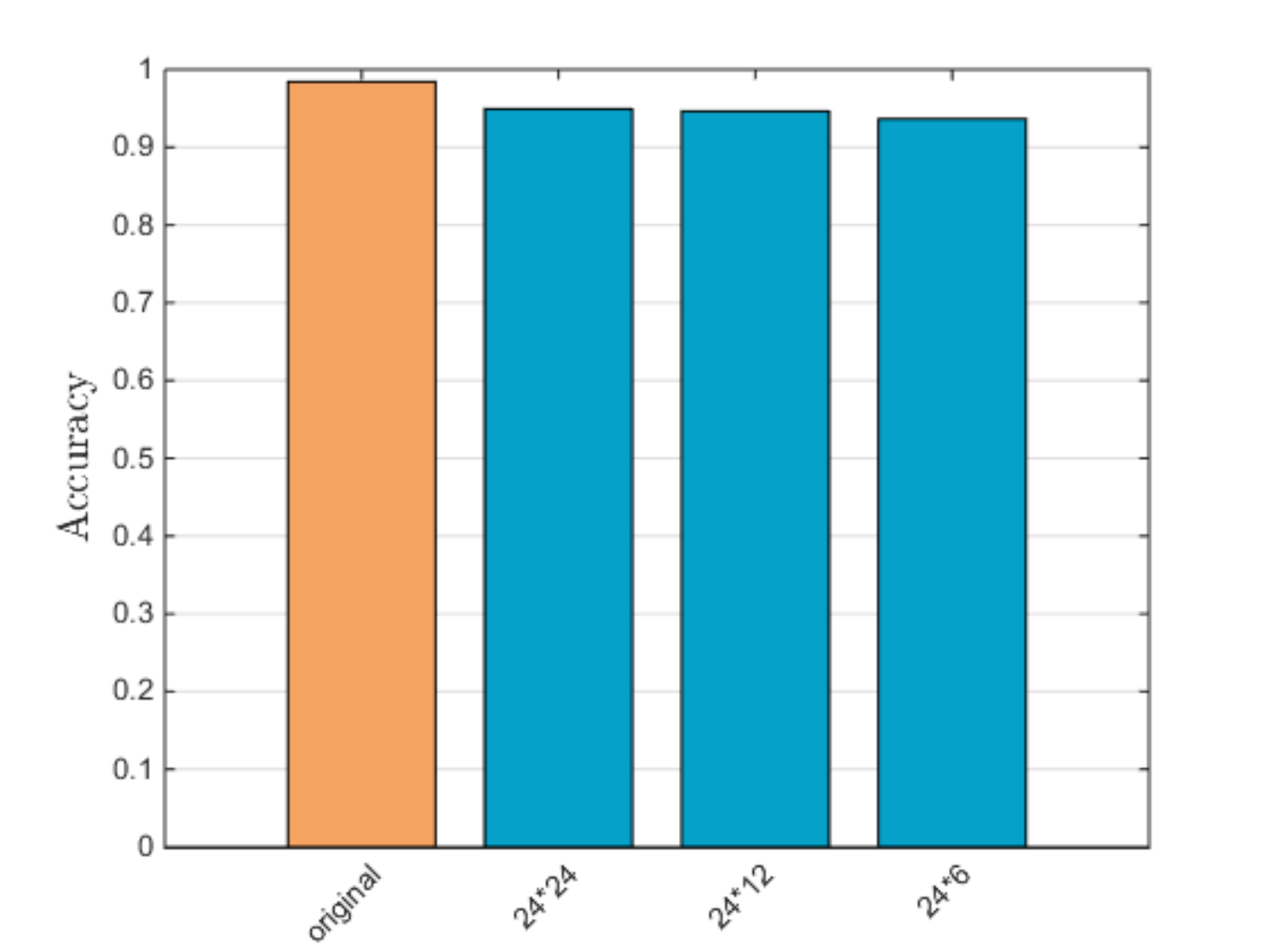}
\label{4-3}
\end{minipage}
}%
\centering
\caption{Accuracy achieved on the CNBC dataset with different $L_b$ on ResNet50}
\label{CNBCRes}
\end{figure}
    
    \item DenseNet121 (Fig. \ref{CNBCDen})
    \begin{enumerate}
        \item $L_b = 96, 48$. The accuracy is stable and beyond 93\% and $W_b$ has minimal impact on the accuracy.
        \item $L_b = 24$. The accuracy is higher than 90\%, and when $W_b = 24$ the accuracy can reach approximately 95\%.
    \end{enumerate}
    
\begin{figure}[!htbp]
\centering

\subfigure[$L_b$ $=$ 96]{
\begin{minipage}[t]{0.30\linewidth}
\centering
\includegraphics[width=\linewidth]{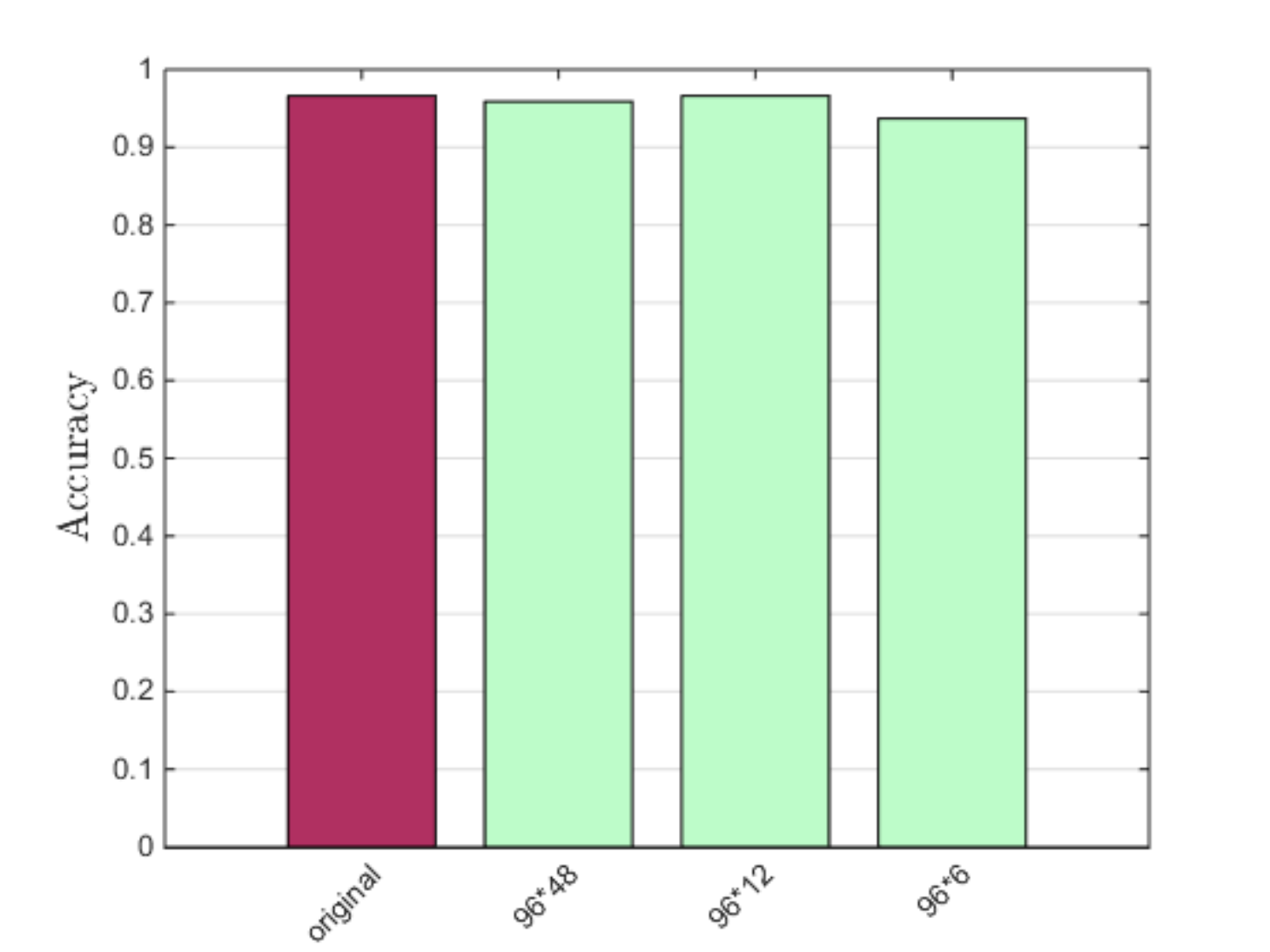}
\label{5-1}
\end{minipage}%
}%
\subfigure[$L_b$ $=$ 48]{
\begin{minipage}[t]{0.30\linewidth}
\centering
\includegraphics[width=\linewidth]{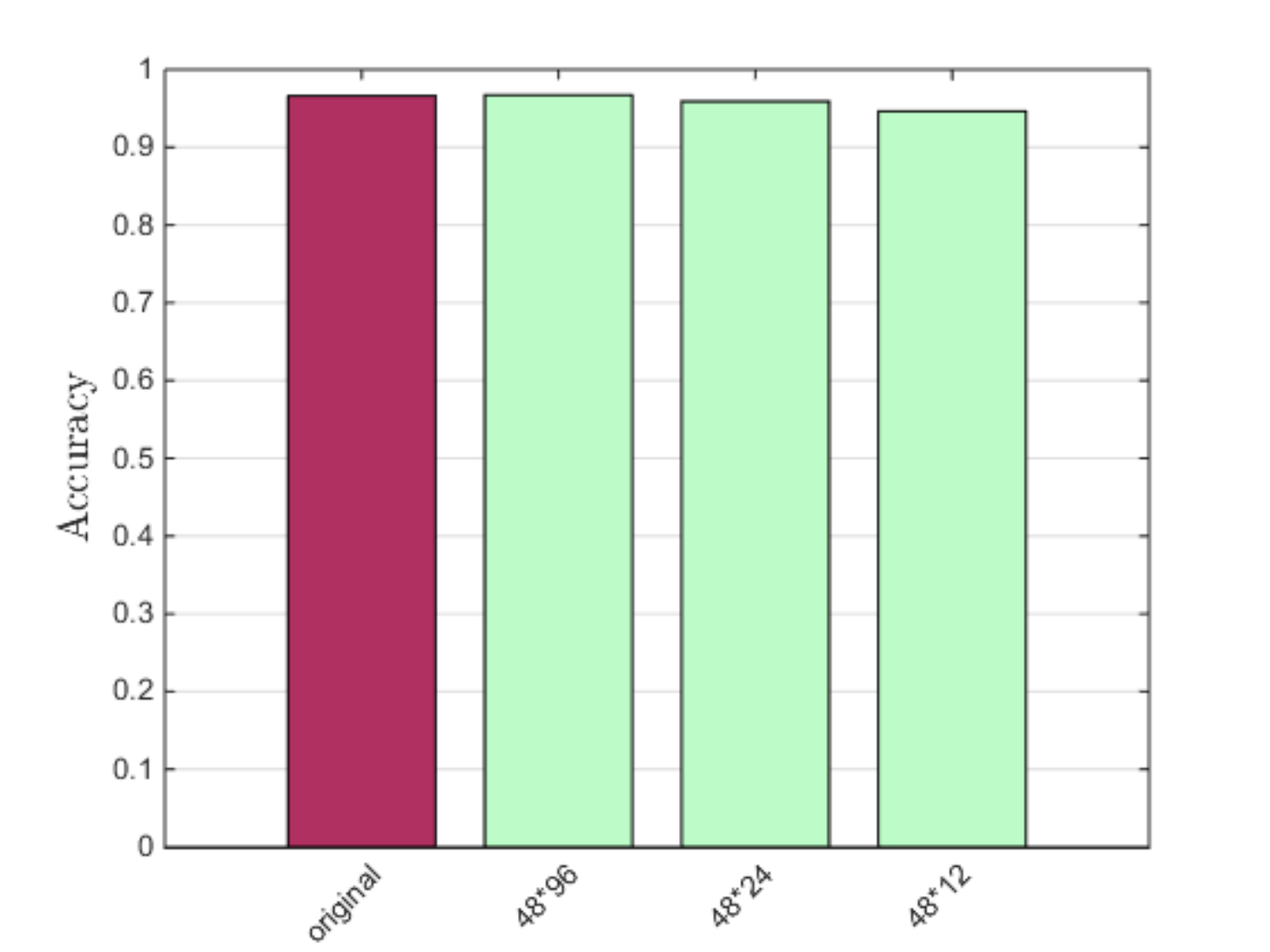}
\label{5-2}
\end{minipage}%
}%
\subfigure[$L_b$ $=$ 24]{
\begin{minipage}[t]{0.30\linewidth}
\centering
\includegraphics[width=\linewidth]{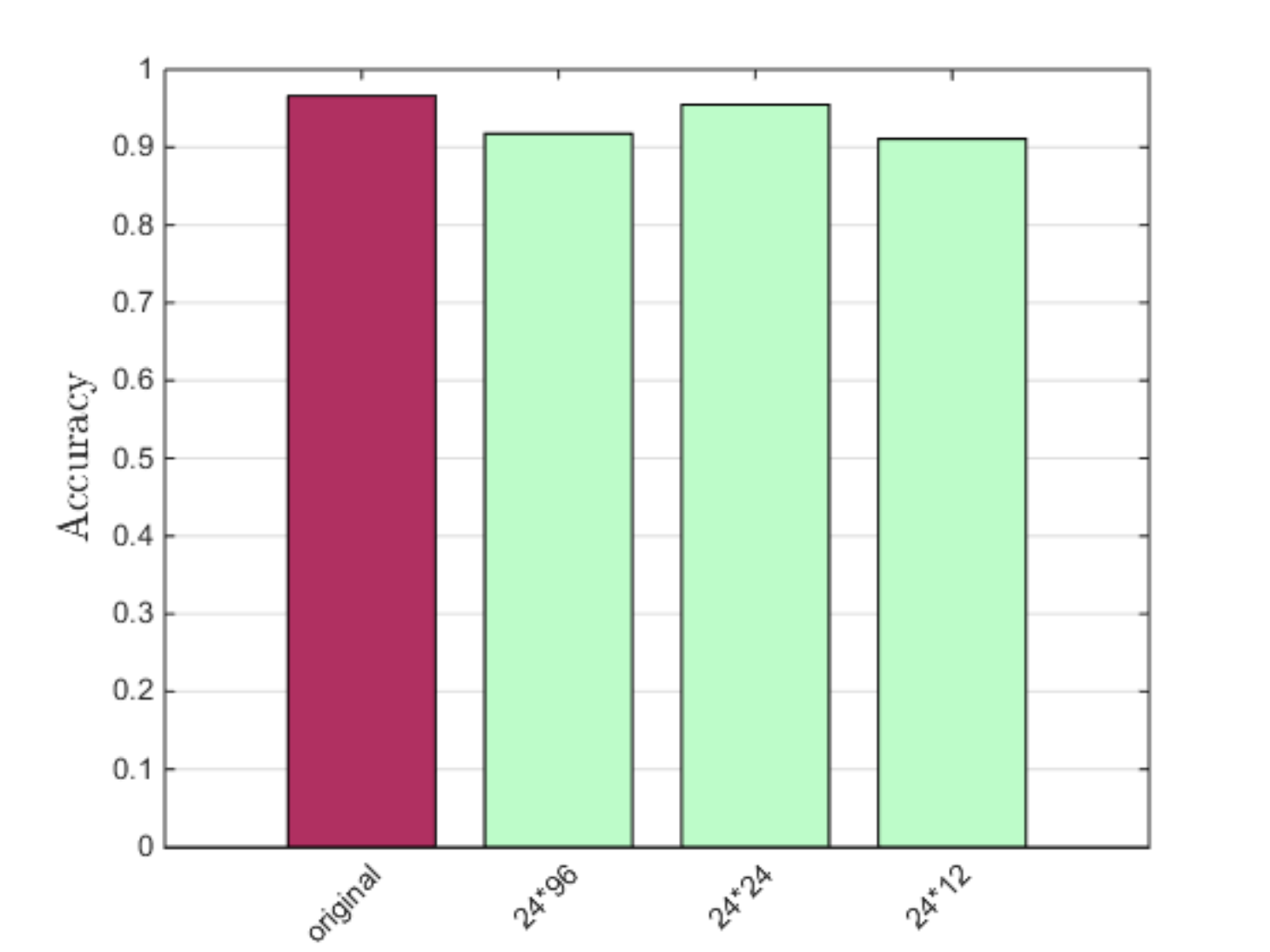}
\label{5-3}
\end{minipage}
}%
\centering
\caption{Accuracy achieved on the CNBC dataset with different $L_b$ on DenseNet121}
\label{CNBCDen}
\end{figure}

\end{itemize}

In summary, for the CNBC face dataset, the accuracy on the testing set of ResNet50 and DenseNet121 models trained by the mixed training set is quite stable and beyond 91\%. Furthermore, $L_b$ and $W_b$ have little influence on the accuracy, while the loss is micro. 

Therefore, on the CNBC dataset our scheme generally achieves high data availability on both ResNet50 and DenseNet121 models.

\subsection{Experiment Summary}

On the basis of the experimental results on the WIKI dataset and the CNBC face dataset, we can see that the pixel block mixing algorithm performs well for both ResNet50 and InceptionV3 models on the WIKI dataset. However, it excels for the VGG16 model on the WIKI dataset, for both ResNet50 and DenseNet121 on the CNBC dataset. 

Besides, we analyze the results more comprehensively according to the results of a single experiment and the results of the whole experiment in Tables \ref{SumWIKI} and \ref{SumCNBC}.

\begin{table}[!htbp]
  \centering
  \caption{Accuracy Overview on the WIKI dataset}
  \label{SumWIKI}
  \begin{tabular}{lcccc}
    \toprule
     & \textbf{Original} & Highest & Lowest & Mean\\ 
    \midrule
    ResNet50 & \textbf{80.40\%} & 80.21\% & 52.79\% & 65.55\% \\
    VGG16 & \textbf{75.13\%} & 79.79\% & 63.49\% & 70.06\%\\
    InceptionV3 & \textbf{84.52\%} & 84.30\% & 52.52\% & 71.63\%\\
    \bottomrule
  \end{tabular}
\end{table}

\begin{table}[!htbp]
  \centering
  \caption{Accuracy Overview on the CNBC dataset}
  \label{SumCNBC}
  \begin{tabular}{lcccc}
    \toprule
     & \textbf{Original} & Highest & Lowest & Mean\\ 
    \midrule
    ResNet50 & \textbf{98.43\%} & 99.01\% & 87.94\% & 95.29\% \\
    DenseNet121 & \textbf{96.62\%} & 98.75\% & 88.78\% & 95.18\%\\
    \bottomrule
  \end{tabular}
\end{table}

On the WIKI dataset, according to the difference between the highest accuracy and the lowest accuracy and the difference between the original accuracy and the mean accuracy, we can see from Table \ref{SumWIKI} that among the models trained by the mixed training set, VGG16 model is more stable than ResNet50 and InceptionV3 models.

For the CNBC face dataset, Table \ref{SumCNBC} shows that the highest accuracy, the lowest and the mean accuracy are relatively steady comparing with the original accuracy on both ResNet50 and DenseNet121 models. Besides, the models trained on the CNBC face dataset perform superior to the models trained on the WIKI dataset.

In summary, the training set mixed by the pixel block mixing algorithm still can be applied to train deep learning models while maintaining the utility. Additionally, we can adjust parameters in the algorithm to balance the relationship between privacy preservation and data availability.

\subsection{Enhancement by Data Augmentation}

\begin{figure}[!htbp]
  \centering
  \includegraphics[width=\linewidth]{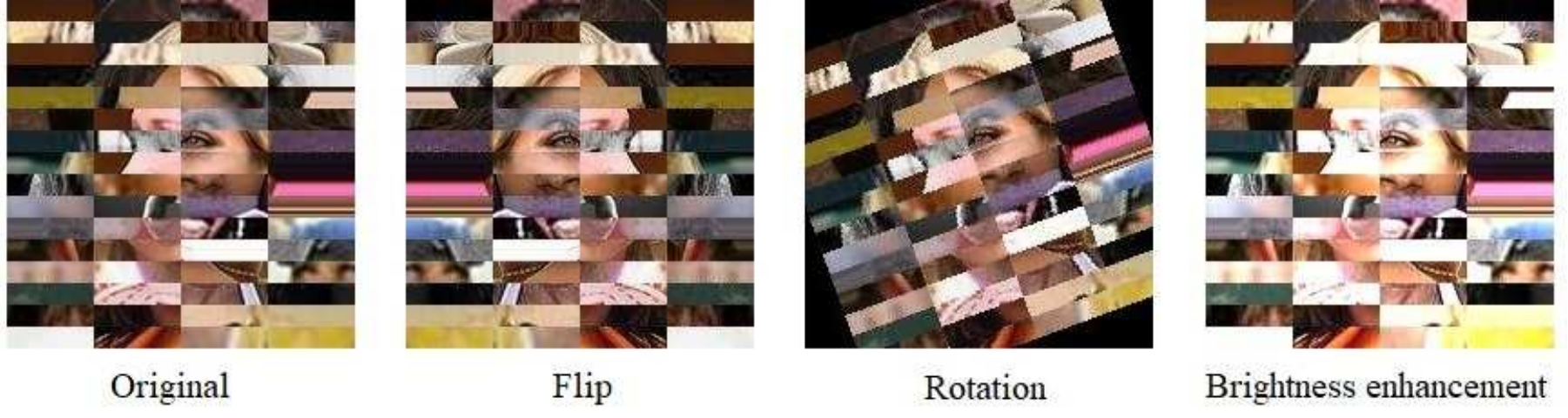}
  \caption{Data augmentation for one image on the WIKI dataset}
  \label{resultenhance}
\end{figure}

In order to improve the accuracy and stabilize the loss, we use data augmentation to enhance our scheme. We use three simple methods for images to implement the data augmentation, which are "flip", "rotation" and "brightness enhancement" respectively. The effect of the image after data augmentation is shown in Fig. \ref{resultenhance}.

\begin{figure}[!htbp]
  \centering
  \includegraphics[width=0.9\linewidth]{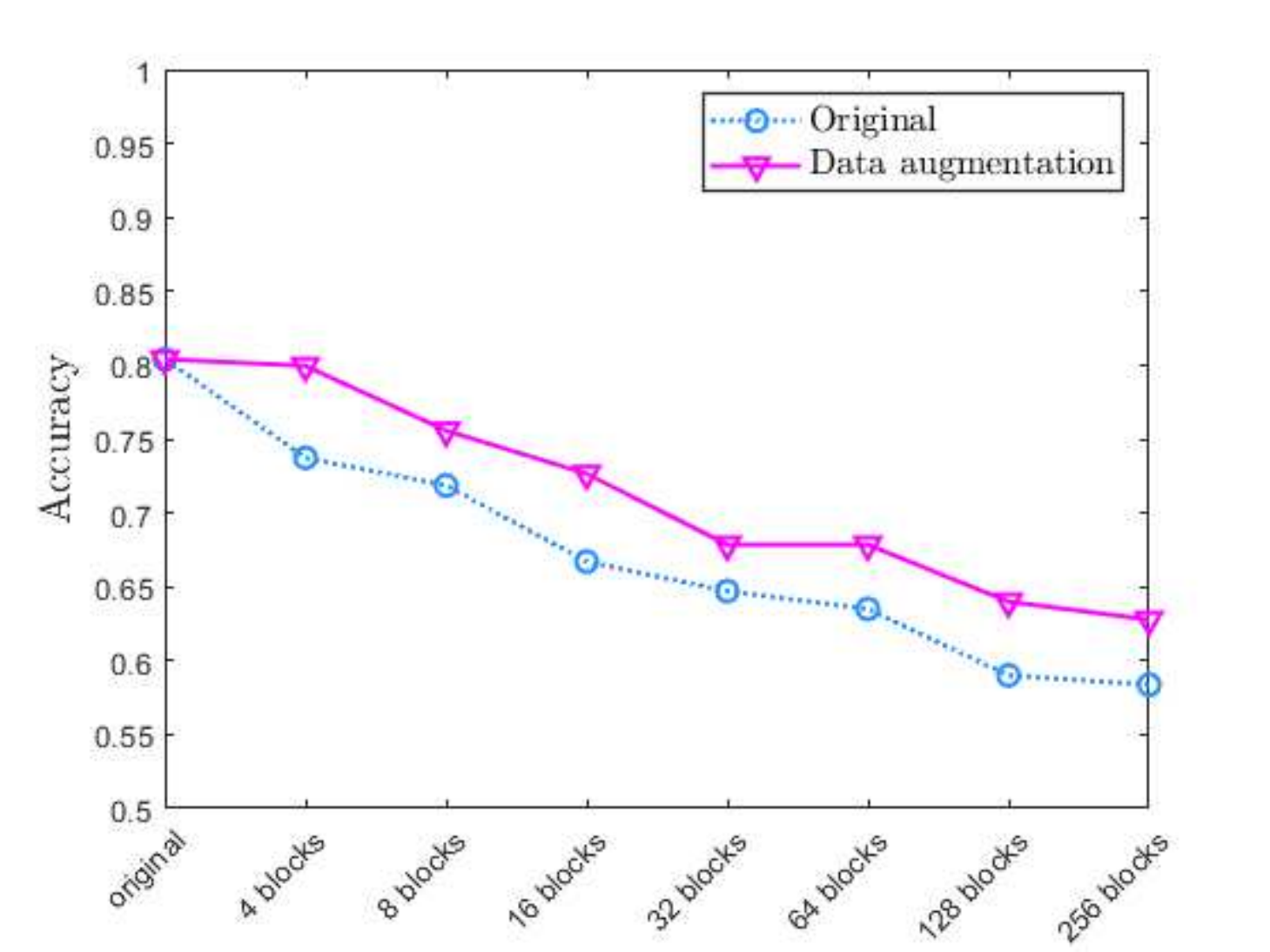}
  \caption{Accuracy of ResNet50 on the WIKI dataset after data augmentation}
  \label{EnhanceAccuracy}
\end{figure}

\begin{figure}[!htbp]
  \centering
  \includegraphics[width=0.9\linewidth]{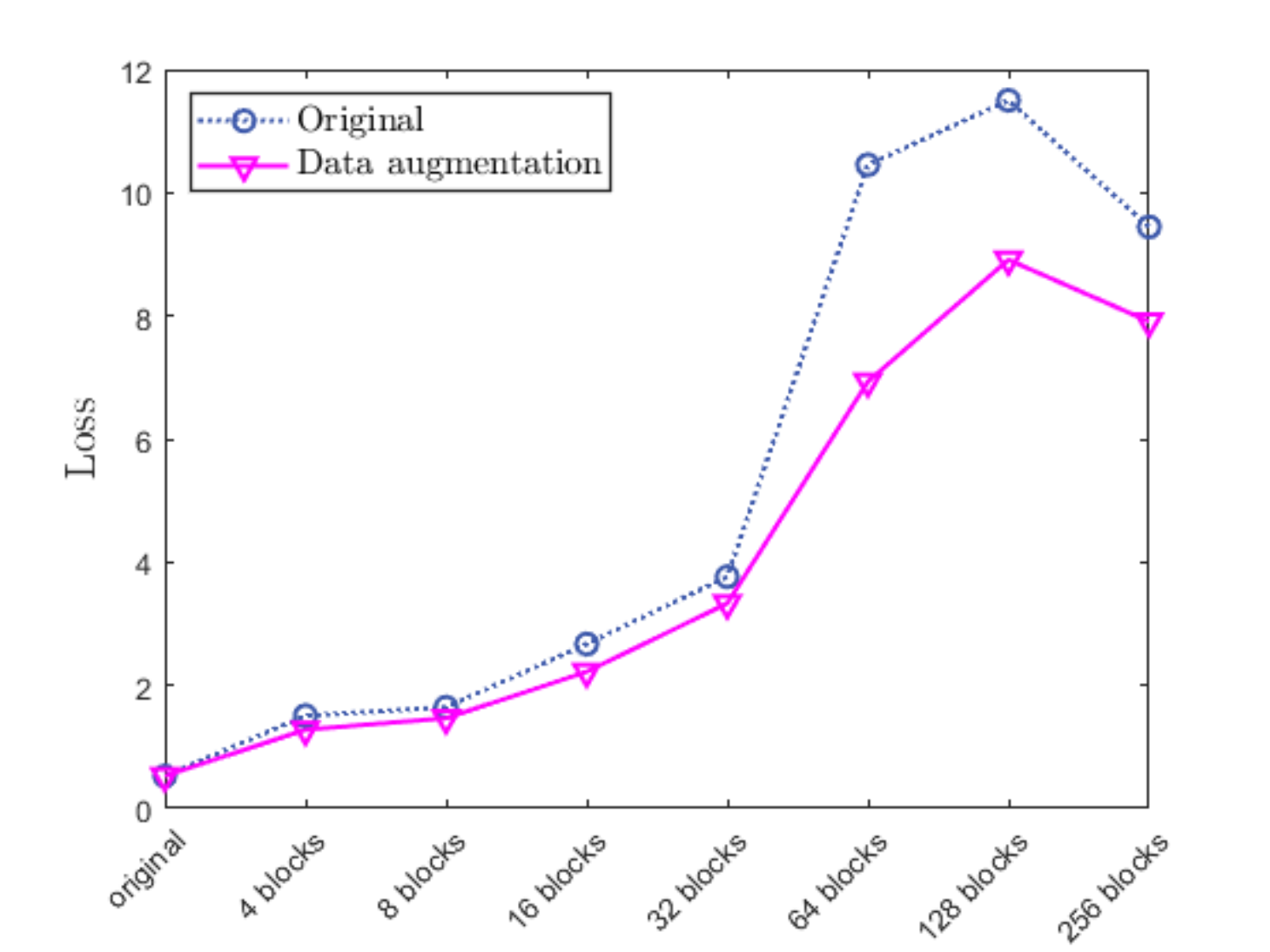}
  \caption{Loss of ResNet50 on the WIKI dataset after data augmentation}
  \label{EnhanceLoss}
\end{figure}

We retest the ResNet50 model on the WIKI dataset with the mixed training set that performs the data augmentation for each image. The results between the former scheme and enhanced scheme are shown in Fig. \ref{EnhanceAccuracy} and Fig. \ref{EnhanceLoss}. From the line charts, obviously, after the application of data augmentation, the accuracy of ResNet50 model on the testing set has improved markedly, and the loss is fewer than the original results too. That also means data augmentation can be applied to the training set mixed by the pixel block mixing algorithm for enhancing the training results.

\section{Scheme Analysis}

\subsection{Security Analysis}
\label{sapbt}

It is critical for the pixel block mixing algorithm to ensure the attackers are not able to restore the original training set through $I_T$ (i.e. guarantee that the personal characteristic information combination of the original dataset can not be reconstructed). In order to restore $I_T$ to $I_S$, the attackers have to restore every category in $I_T$. We assume there are $p$ types in $I_T$, each type is $C_i$, and $N_{C_{i}}$ images in each corresponding type.

There are three situations that will happen to the attackers who are trying to restore a category in $I_T$ to the original category in $I_S$:

\begin{itemize}
    \item At least one repeated pixel block exists in $C_i$.
    \item At least one original pixel block is erased in $C_i$.
    \item Every original pixel block exists in $C_i$.
\end{itemize}

Further analysis of the above situations shows that the first two situations are actually the same. Because the whole space volume in each category is fixed, when one repeated pixel block exists that also means one original pixel block has been erased. According to the above analysis, the attackers can not restore the original category in this situation. 

Therefore, the only situation the attacker may be able to restore the category in $I_T$ we need to consider is when every original pixel block exists. We divide the images in one category into $N_b$ groups, each group includes all pixel blocks in the same position in each image. 

If the attackers try to restore every image in one category (e.g. $C_1$), the probability is:

\begin{equation}\label{security1}
\frac{1}{{(N_b!)}^{N_{C_{1}}}} = \frac{1}{{N_b}^{N_{C_{1}}}} \cdot \frac{1}{{(N_b-1)}^{N_{C_{1}}}} \cdot \cdot \cdot \frac{1}{{2}^{N_{C_{1}}}} \cdot \frac{1}{{1}^{N_{C_{1}}}}
\end{equation}

When every original pixel block exists in every category in $I_T$, if the attackers intend to restore the whole $I_T$ to $I_S$ the probability is:

\begin{equation}\label{security2}
\prod_{i=1}^{p}\frac{1}{{(N_b!)}^{N_{C_{i}}}}
\end{equation}

We can see from Eq.(\ref{security1}) and Eq.(\ref{security2}) that if there are enough images in each category, even if every original pixel block exists, it is nearly impossible for the attackers to restore even a category in the training set. Thus, our scheme is secure and able to defend against restoring attacks.

\subsection{Privacy Preservation Analysis}

We utilize the structural similarity index measure (SSIM) \cite{wang2004image} to calculate the privacy preservation effectiveness of the training set mixed by the pixel block mixing algorithm with different $N_b$ on the WIKI dataset and the CNBC face dataset.

A smaller value of SSIM indicates that the original training set and the mixed training set are less similar, i.e., the better privacy preservation. The results are demonstrated in Table \ref{SSIM}. 

Overall, it is evident that SSIM decreases with the $N_b$ raises, which means the larger $N_b$, the greater the relative effectiveness of privacy preservation in general. However, the special trends in the results also need to be discussed. The specific discussion will be shown in Section \ref{sec:d}.

\begin{table*}[!htbp]
  \centering
  \caption{Privacy Preservation Effectiveness (SSIM)}
  \label{SSIM}
  \begin{tabular}{c|ccccccc}
    \toprule
    \diagbox{dataset}{$N_b$} & 4 & 8 & 16 & 32 & 64 & 128 & 256\\ 
    \midrule
    \textbf{WIKI} 
 & 0.35640
 & 0.34625
 & 0.32716
 & 0.29683
 & 0.28486
 & 0.24846
 & 0.26440
 \\
    \midrule
    \textbf{CNBC}
 & 0.62312
 & 0.61694
 & 0.60893
 & 0.59498
 & 0.58072
 & 0.55228
 & 0.52728
\\
    \bottomrule
  \end{tabular}
\end{table*}

\subsection{Performance Analysis}

We test the performance of the pixel block mixing algorithm with different $N_{I_{S}}$ and $N_b$. Our experimental environment is displayed in Table \ref{con}. 

\begin{table}[!htbp]
\centering
\caption{Computer Configuration}
\label{con}
\begin{tabular}{|l|l|}
\hline
Processor (CPU) & Intel Core i5-8265U\\
\hline
Memory & 8 GB RAM\\
\hline
Operating System & Microsoft Windows 10 x64\\
\hline
\end{tabular}
\end{table}

\begin{figure}[!htbp]
  \centering
  \includegraphics[width=0.9\linewidth]{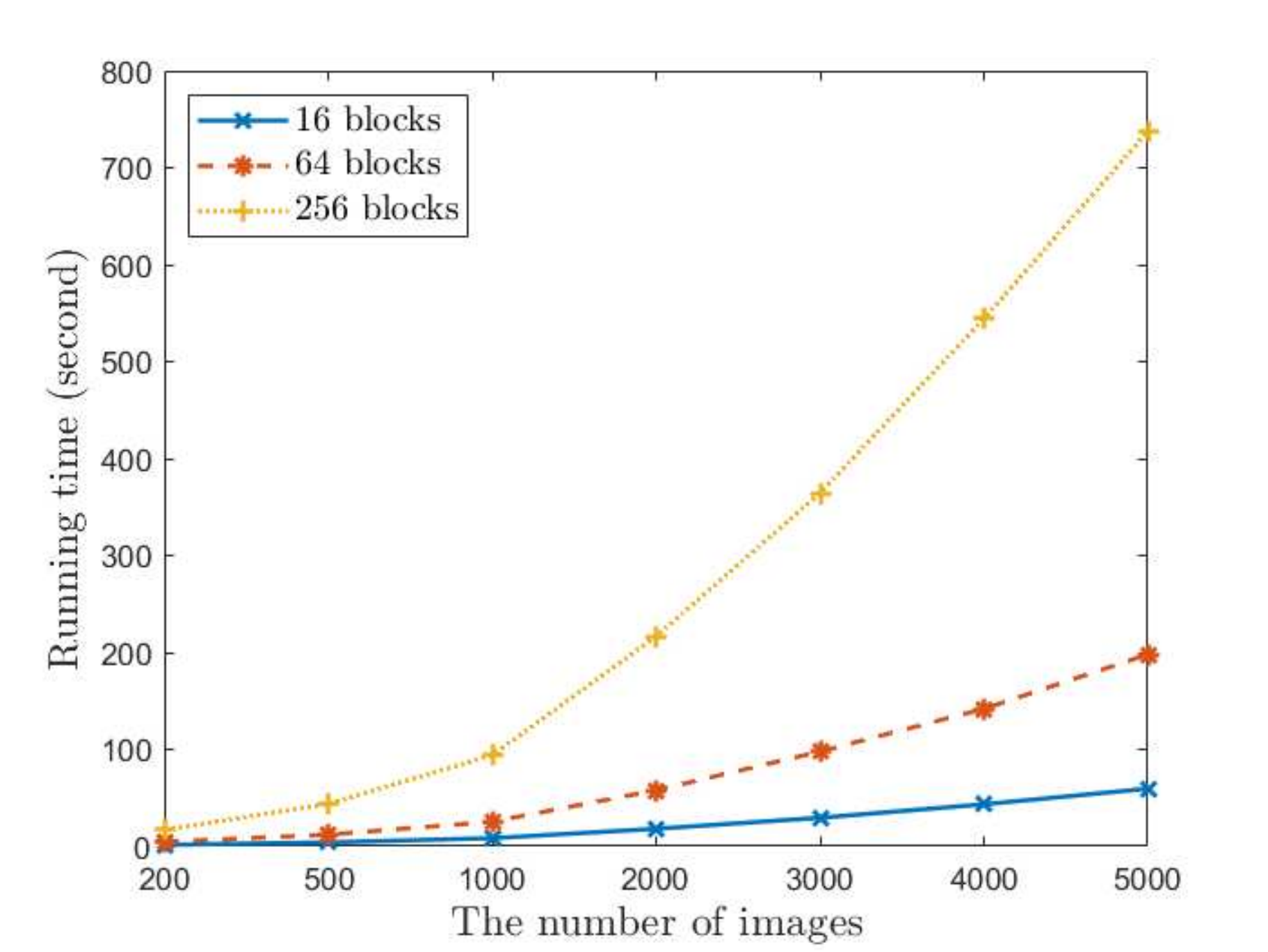}
  \caption{The running time with different $N_{I_{S}}$}
  \label{RTI}
\end{figure}

\begin{figure}[!htbp]
  \centering
  \includegraphics[width=0.9\linewidth]{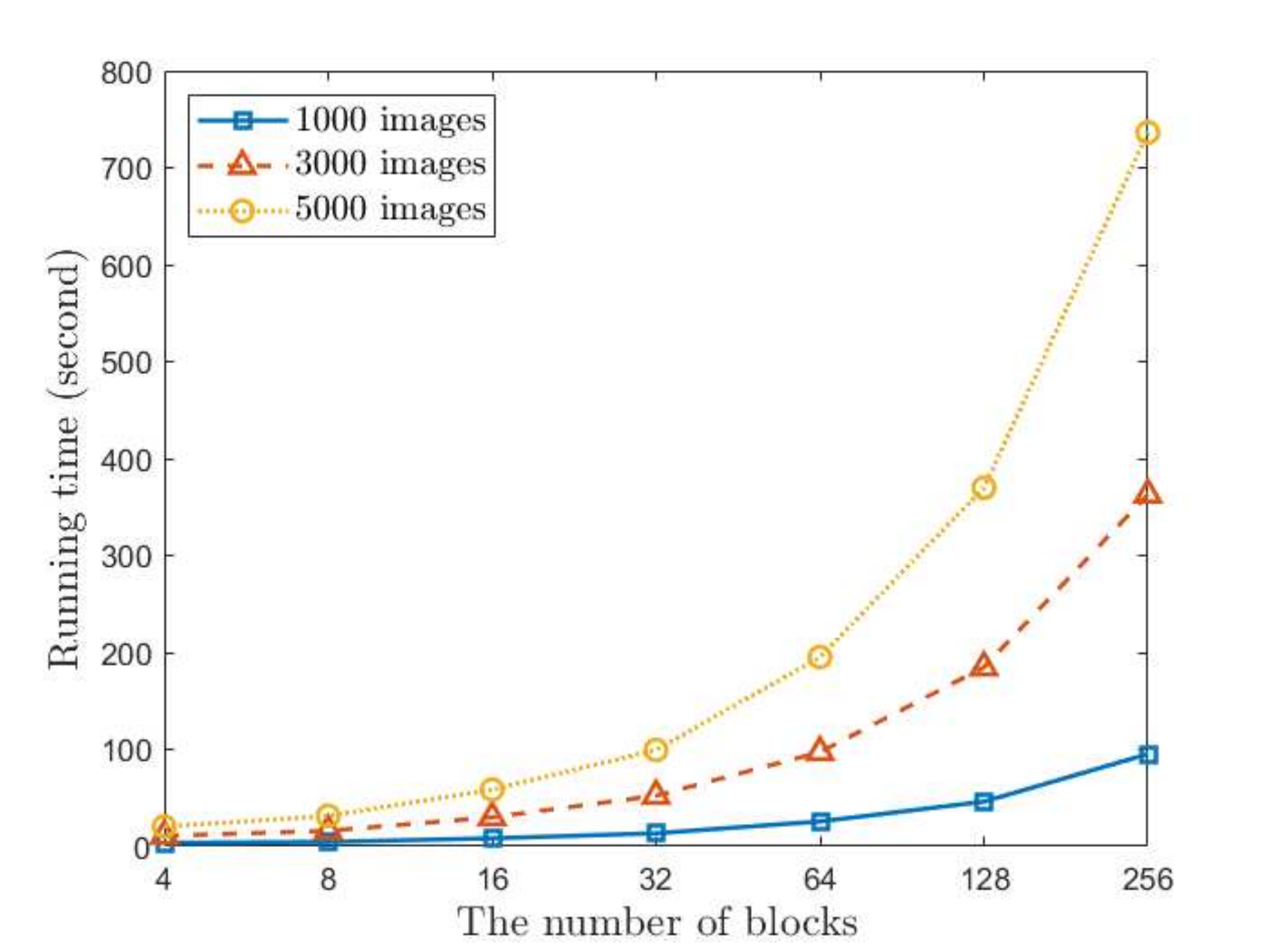}
  \caption{The running time with different $N_b$}
  \label{RTB}
\end{figure}

The results of the performance testing are shown in Fig. \ref{RTI} and Fig. \ref{RTB}. The above figures illustrate that when $N_b$ is fixed, the trend of running time with increasing $N_{I_{S}}$ is nearly a linear image; while when $N_{I_{S}}$ is fixed, the slope of the image raises more as the $N_b$ increases. 

To sum up,  that $N_b$ is a decisive factor in the efficiency of the pixel block mixing algorithm. Besides, when $N_b$ is 128 and 256 and $N_{I_{S}}$ is 5000, the pixel block mixing algorithm can process approximately 13 and 6 images respectively per second in under the configuration of our computer, which also demonstrates our scheme is quite efficient.

\section{Discussion}
\label{sec:d}

\subsection{Accuracy Trend on the WIKI and CNBC datasets}
From Fig.\ref{WIKItrand} we can see that for VGG16 of InceptionV3, there is a significant drop in the accuracy for $N_b = 4$ followed by a considerable increase for $N_b = 8$. A similar trend occurs for $N_b = 128$ and $N_b = 256$. Besides, Fig.\ref{CNBCtrand} shows the pattern of increases and decreases for different $N_b$.

For the accuracy trend on the WIKI and CNBC datasets, we think there are two reasons that might cause the above phenomena:
\begin{itemize}
    \item There are several critical values that exist. If $N_b$ does not reach the critical value, the accuracy rate will fluctuate within a certain range. If the critical value is reached, it will decrease as $N_b$ increases until the next critical value is reached. After reaching the new critical value, the accuracy rate will fluctuate within a certain range, and so on.
    \item The characteristics and robustness of deep neural network models are different according to their different structures. This might result in different trends between the models in the same dataset. Therefore, some deep neural network models might exist one or more critical values in this case, but others not.
\end{itemize}

\subsection{Loss Trend on the WIKI dataset}
In Fig.\ref{WIKIloss}, there is a significant drop in the observed loss from $N_b = 128$ to $N_b = 256$. Except this special trend, the loss increases with the increasing $N_b$.

To explain this, we refer the values in Table \ref{SSIM}. We can see that the SSIM declines when $N_b$ increases, until ${N_b = 256}$. From ${N_b = 128}$ to ${N_b = 256}$, the SSIM goes up (i.e. 0.24846 to 0.26440). This trend might illustrate the special loss trend in Fig.\ref{WIKIloss}.

\subsection{The Differences on the WIKI and CNBC datasets}
Fig.\ref{WIKItrand} and Fig.\ref{CNBCtrand} display that the loss drops less on the CNBC datasets with $N_b$ climbs. Considering the impact of different deep neural network structures, we focus on the ResNet50 model. 

For the ResNet50 model on the CNBC dataset, the accuracy fluctuates up to around 5\%. However, on the WIKI dataset, the accuracy of the ResNet50 model fluctuates about 20\% compared with the accuracy on the original training set when ${N_b = 256}$.

\begin{figure}[!htbp]
  \centering
  \includegraphics[width=0.75\linewidth]{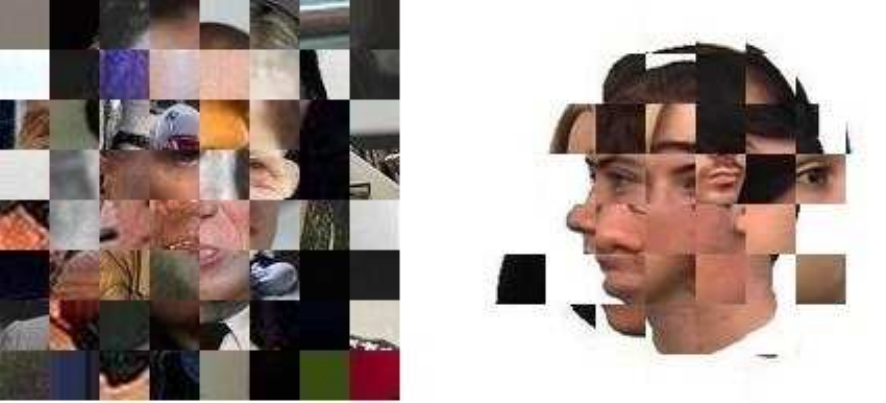}
  \caption{The differences between the WIKI (left) and CNBC (right) datasets}
  \label{DIF}
\end{figure}

We think the reason is that the CNBC dataset is "clearer" than the WIKI dataset. That means, there are fewer interference features in images in the CNBC dataset than the images in the WIKI dataset, see Fig.\ref{DIF}. Consequently, we could also speculate that our solution will perform better on the datasets similar to the CNBC dataset.

\section{Conclusion}
\label{sec:5}
We proposed a lightweight and efficient scheme to preserve privacy in deep learning image classification, based on our pixel block mixing algorithm. Using both WIKI and CNBC face datasets, we evaluated the performance and security of our proposed approach. For example, the experimental results showed that the pixel block mixing algorithm achieves good performance for the VGG16 model on the WIKI dataset, and for both ResNet50 and DenseNet121 models on the CNBC face dataset. Furthermore, we also observed that data augmentation can enhance the accuracy and decrease the loss for the trained ResNet50 model on the WIKI dataset, in comparison to the original results (without using data augmentation). We also observed from the mixed training set, it is nearly impossible for the attackers to restore it to the original training set. In addition, we also observed the privacy preservation's effectiveness improves when $N_b$ increases. The performance evaluation of the pixel block mixing algorithm also showed that our scheme is highly efficient, and $N_b$ is a key factor in deciding efficiency. Therefore, we can adjust the balance between privacy preservation and efficiency by changing $N_b$. We also remark that our pixel block mixing algorithm is simple and has adjustable parameter settings; thus, it can be easily customized to achieve different balance levels between privacy preservation and accuracy flexibly according to the actual needs.

No approach is perfect. There are a number of potential future research agendas. First, what is the optimal key parameter setting (e.g.,  $N_s$ and $N_t$) in the pixel block mixing algorithm? If we divide the images in the training set unevenly then mixing, can we achieve better results?


%



\section*{Acknowledgment}
The research was financially supported by National Natural Science Foundation of China (No. 61972366), the Provincial Key Research and Development Program of Hubei (No. 2020BAB105), the Henan Key Laboratory of Network Cryptography Technology (No. LNCT2020-A01), the Foundation of Guangxi Key Laboratory of Cryptography and Information Security (No. GCIS201913), the Foundation of Key Laboratory of Network Assessment Technology, Chinese Academy of Sciences (No. KFKT2019-003), Major Scientific and Technological Special Project of Guizhou Province (No. 20183001), and the Foundation of Guizhou Provincial Key Laboratory of Public Big Data (No. 2018BDKFJJ009, No. 2019BDKFJJ003, No. 2019BDKFJJ011). K.-K.R. Choo was funded only by the Cloud Technology Endowed Professorship.


\ifCLASSOPTIONcaptionsoff
  \newpage
\fi




\bibliographystyle{IEEEtran}
\bibliography{ref}

\end{document}